\documentclass[twocolumn]{article}
\usepackage{authblk}

\usepackage{times}
\usepackage{amsmath,amssymb}
\usepackage{amsfonts}
\usepackage{mathtools}
\usepackage{sty/wdel}
\usepackage{sty/notations}
\usepackage{sty/dm-colors}

\usepackage{cite}
\usepackage{algorithm}
\usepackage{algorithmicx}
\usepackage{algpseudocode}
\usepackage{graphicx}
\usepackage{textcomp}
\usepackage{xcolor}
\usepackage{bm}
\usepackage[caption=false]{subfig} 

\usepackage[whole]{bxcjkjatype} % For Japanese output with pdfLaTeX (kept for compatibility)
\allowdisplaybreaks[1]

\usepackage{xcolor}

\newcommand{\set}[1]{{\left\{{#1}\right\}}}

\newtheorem{definition}{Definition}

\usepackage{hyperref}
\hypersetup{
  colorlinks,
  citecolor=dmblue500,
  linkcolor=dmblue500,
  urlcolor=dmblue500
}

\usepackage[capitalize,noabbrev,nameinlink]{cleveref}

\crefname{algorithm}{Algorithm}{Algorithms}
\crefname{assumption}{Assumption}{Assumptions}
\crefname{corollary}{Corollary}{Corollaries}
\crefname{definition}{Definition}{Definitions}
\crefname{equation}{Equation}{Equations}
\crefname{example}{Example}{Examples}
\crefname{figure}{Figure}{Figures}
\crefname{lemma}{Lemma}{Lemmas}
\crefname{proposition}{Proposition}{Propositions}
\crefname{remark}{Remark}{Remarks}
\crefname{table}{Table}{Tables}
\crefname{theorem}{Theorem}{Theorems}

\Crefname{algorithm}{Algorithm}{Algorithms}
\Crefname{assumption}{Assumption}{Assumptions}
\Crefname{corollary}{Corollary}{Corollaries}
\Crefname{definition}{Definition}{Definitions}
\Crefname{equation}{Equation}{Equations}
\Crefname{example}{Example}{Examples}
\Crefname{figure}{Figure}{Figures}
\Crefname{lemma}{Lemma}{Lemmas}
\Crefname{proposition}{Proposition}{Propositions}
\Crefname{remark}{Remark}{Remarks}
\Crefname{table}{Table}{Tables}
\Crefname{theorem}{Theorem}{Theorems}

\usepackage{autonum}
\makeatletter
\autonum@generatePatchedReferenceCSL{Cref}
\makeatother

\usepackage{multirow}
\usepackage{booktabs}

\title{Symmetry-Breaking in Multi-Agent Navigation: Winding Number-Aware MPC with a Learned Topological Strategy}

\author[*, 1]{%
	Tomoki Nakao%
}
\author[*, 2]{%
Kazumi Kasaura
}%
\author[2]{%
Tadashi Kozuno
}%
\affil[1]{Graduate School of Informatics, Kyoto University, Kyoto, Japan.
This work was done while he was a research intern at OMRON SINIC X Corporation. \texttt{nakao.tomoki.54e@st.kyoto-u.ac.jp}}
\affil[2]{OMRON SINIC X Corporation, 5-24-5, Hongo, Bunkyo-ku, Tokyo, Japan. \texttt{\{kazumi.kasaura, tadashi.kozuno\}@sinicx.com}}
\affil[*]{Equal contribution.}
\date{}

\begin{document}

\maketitle

%%%%%%%%%%%%%%%%%%%%%%%%%%%%%%%%%%%%%%%%%%%%%%%%%%%%%%%%%%%%%%%%%%%%%%%%%%%%%%%%

\begin{abstract}
In decentralized multi-agent navigation, agents that independently compute their controls without communicating goals or intentions can fall into symmetry-induced deadlocks because each agent must decide how to pass others. We study this problem under the assumption that each agent has access to the current observable states of other agents, including their positions, velocities, and radii, while their goals, intentions, and future trajectories remain unobserved. 
To address this problem, we propose WNumMPC, a hierarchical navigation method that quantifies cooperative symmetry-breaking strategies via a topological invariant, the winding number, and learns such strategies through reinforcement learning. The learning-based Planner outputs continuous-valued signed target winding numbers and dynamic importance weights to prioritize critical interactions in dense crossings.
Then, the model-based Controller generates collision-free and efficient motions based on the strategy and weights provided by the Planner. 
Simulation and real-world robot experiments indicate that WNumMPC effectively avoids deadlocks and collisions and achieves better performance than the baselines, particularly in dense and symmetry-prone scenarios. These experiments also suggest that explicitly leveraging winding numbers yields robust sim-to-real transfer with minimal performance degradation.
The code for the experiments is available at \url{https://github.com/omron-sinicx/WNumMPC}.
\end{abstract}

%%%%%%%%%%%%%%%%%%%%%%%%%%%%%%%%%%%%%%%%%%%%%%%%%%%%%%%%%%%%%%%%%%%%%%%%%%%%%%%%
%%%%%%%%%%%%%%%%%%%%%%%%%%%%%%%%%%%%%%%%%%%%%%%%%%%%%%%%%%%%%%%%%%%%%%%%%%%%%%%%

\section{Introduction} \label{sec:introduction}
The multi-agent navigation problem considers multiple agents moving in a shared space to reach their respective goals while avoiding collisions. This problem is central to many robotics applications such as warehouse automation and traffic management, and has been studied from various perspectives. 
% A straightforward approach is centralized navigation; however, it requires inter-agent communication and scales poorly computationally for large systems~\cite{central_mpc, MPPI_paralell}, making deployment in real-world settings challenging. 
% A straightforward approach is centralized navigation; however, its computational cost increases rapidly with the number of agents and becomes computationally demanding for large systems~\cite{central_mpc, MPPI_paralell}, making deployment in real-world settings challenging.
A straightforward approach is centralized navigation; however, it requires explicit inter-agent communication, and its computational cost often increases rapidly with the number of agents~\cite{camponogara2002distributed, MPPI_paralell}, making deployment in real-world settings challenging.
In contrast, distributed navigation does not require explicit communication and is scalable with respect to the number of agents; thus, it has received significant attention~\cite{CADRL, GA3C, SA-CADRL}. In distributed settings, since others' intentions (e.g., goals) are unobservable, agents must plan cooperatively, going beyond purely local collision avoidance.

A key challenge arises from \emph{symmetry} among agents in the absence of explicit communication or priority. For example, when two agents approach each other, they may fail to decide on a passing side and become stuck~\cite{robot_dance}. Such symmetry-induced deadlocks are a major obstacle for distributed navigation; the system needs to recognize interactions and \emph{break symmetry} cooperatively.
%Incorporating the \emph{topological} relation between trajectories is effective for this purpose~\cite{Mav21, Mav23}.
On the other hand, when multiple agents pass each other on a plane, abstracting their passing patterns using topological features is effective in crowd navigation~\cite{Mav21, Mav23}.
In particular, the winding number provides a topological invariant that abstracts the essence of passing, enabling a quantitative description of how trajectories wind around each other.
However, the symmetry issue still exists in navigation with winding numbers, and since rule-based methods have difficulty in breaking symmetry, a flexible learning-based approach offers a more promising direction.

\begin{figure}[t]
    \centering
    \includegraphics[width=1.0\linewidth]{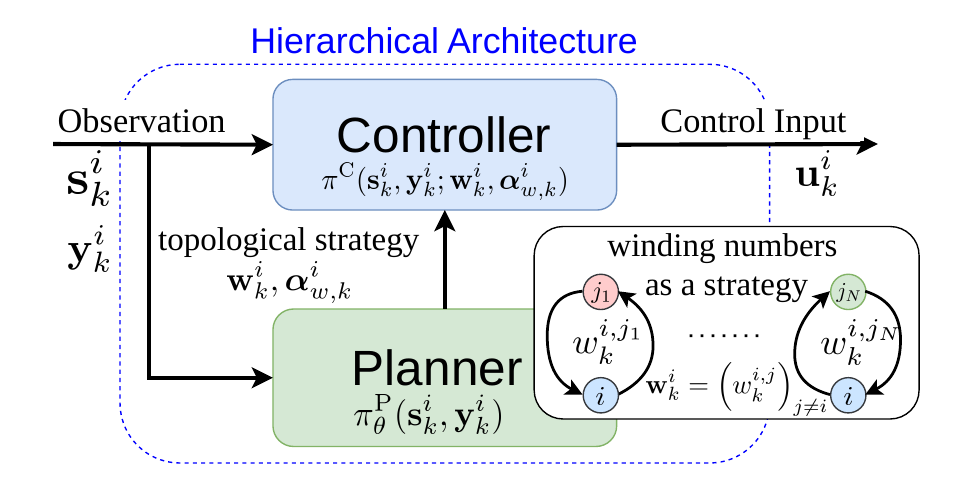}
    \caption{
        % The proposed hierarchical architecture for cooperative symmetry breaking.
        % The learning-based Planner generates a high-level topological strategy to break symmetry,
        % which the model-based Controller then executes reliably.
        Overview of the proposed hierarchical architecture for cooperative symmetry breaking. 
        Each agent independently runs a learning-based Planner to generate a high-level topological strategy for symmetry breaking 
        and a model-based Controller to execute the strategy reliably.
    }
    \label{fig:framework}
\end{figure}

In this work, we address symmetry-breaking for distributed navigation by proposing a winding-number-aware navigation algorithm. The framework of our proposed method is illustrated in \cref{fig:framework}. Our method adopts a hierarchical policy with a learning-based Planner that determines a global cooperative strategy and a model-based Controller that executes local collision avoidance.
The Planner learns both a topological cooperative strategy represented as a winding number and dynamic weights indicating which agents to prioritize for coordination.
The Controller then executes local control to follow the plan.

%\subsection*{Contributions} \label{sec:contributions}
% In this work, we address the fundamental challenge of deadlocks caused by positional symmetry in distributed navigation, and propose a new hierarchical navigation method that learns cooperative strategies for symmetry breaking via reinforcement learning. The framework of our proposed method is illustrated in Fig.~\ref{fig:framework}. The contributions of this paper are as follows:
%We propose a new hierarchical navigation method that learns cooperative strategies for symmetry breaking via reinforcement learning. 
% The framework of our proposed method is illustrated in \cref{fig:framework}.
The contributions of our method are as follows.
First, we propose a hierarchical framework that unifies strategy planning by the Planner and reliable execution by the Controller, using the notion of winding numbers.
% Second, we develop a multi-agent reinforcement learning algorithm to train the Planner to output topological cooperative strategies for symmetry breaking.
Second, we train the Planner via multi-agent reinforcement learning to output topological cooperative strategies for symmetry breaking. 

\begin{figure}[t]
    \centering
    \includegraphics[width=0.7\linewidth]{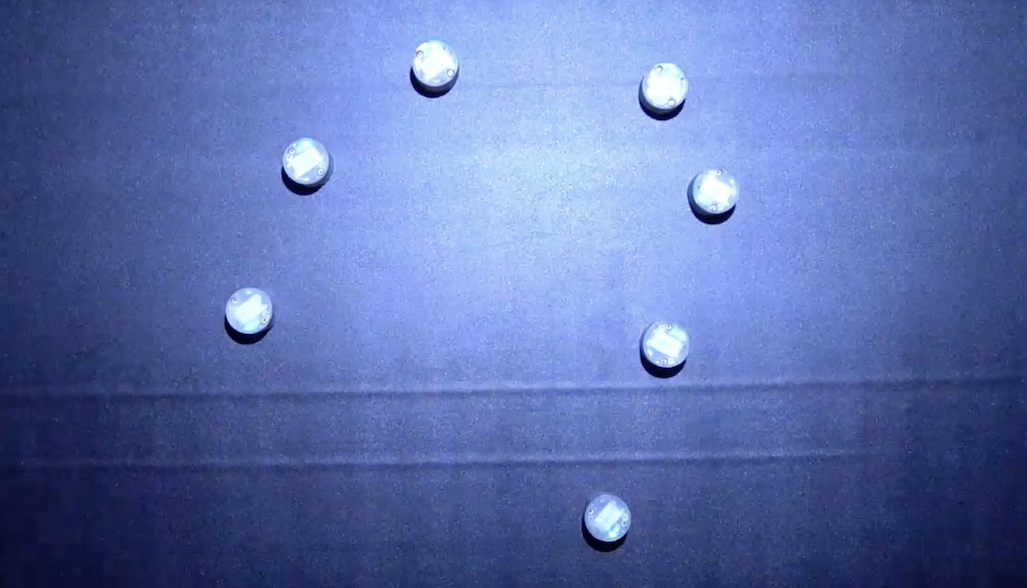}
    \caption{
        % Seven small two-wheeled robots (“maru”~\cite{ichihashi2024swarm}) moving on a tabletop. 
        Seven small two-wheeled tabletop robots (\emph{maru}~\cite{ichihashi2024swarm}) navigating efficiently by cooperatively breaking positional symmetry.
        % By cooperatively breaking positional symmetry, robots achieve efficient navigation.
    }
    \label{fig:wnum_clip}
\end{figure}

We validate the effectiveness of our method through extensive simulation and real-world experiments with tabletop robots named \emph{maru} (\cref{fig:wnum_clip}). By comparing against established baselines, we quantitatively demonstrate the advantages of employing a hierarchical architecture and learning topological cooperative strategies.

%%%%%%%%%%%%%%%%%%%%%%%%%%%%%%%%%%%%%%%%%%%%%%%%%%%%%%%%%%%%%%%%%%%%%%%%%%%%%%%%
%%%%%%%%%%%%%%%%%%%%%%%%%%%%%%%%%%%%%%%%%%%%%%%%%%%%%%%%%%%%%%%%%%%%%%%%%%%%%%%%

\section{Related Work} \label{sec:related_works}
Distributed multi-agent navigation is a challenging problem and an actively studied area. 
Reactive methods such as \cite{SF, RVO, ORCA} achieve collision avoidance by considering one-step interactions based on geometric or physical relations. 
These approaches offer excellent computational efficiency; however, since they do not account for the temporal evolution of the states of surrounding agents,
they are known to result in short-sighted behaviors~\cite{CADRL}. 
To address this limitation, trajectory-based methods have been proposed that plan navigation based on long-horizon predictions. For example, \cite{dMPPI} estimates the local goals of surrounding agents using constant-velocity assumptions and performs model-based planning over a long horizon. However, these approaches rely entirely on hand-crafted rules or cost functions, which limits their ability to achieve cooperative behaviors and generalize to dense environments with complex interactions. In particular, when the spatial relations among agents are symmetric, agents may give way to each other, leading to deadlocks.
To overcome the limitations in cooperativeness and generalization, many learning-based methods have been explored. Approaches such as \cite{CADRL, SA-CADRL, GA3C} learn policies that account for inter-agent interactions to achieve collision avoidance. CADRL and its variants~\cite{CADRL, SA-CADRL, GA3C} employ reinforcement learning to refine policies after pretraining with trajectories from ORCA~\cite{ORCA}. However, this framework inherits the limitations of ORCA~\cite{DS-RNN}.

A fundamental difficulty common to the above methods lies in cooperative decision-making that requires breaking symmetry. 
A promising direction is to focus on the topological relations among trajectories. Several works have explored topological concepts to represent agent interactions beyond simple geometric relations. 
For instance, some works have utilized braid theory~\cite{mavrogiannis2019multi,mavrogiannis2022implicit} to capture finer topological characteristics for systems with more than two agents. 
% Another hierarchical approach~\cite{cao2019dynamic} computes topological features by enumerating crossings with edges in a graph connecting other agents. 
Meanwhile, \cite{Mav21, Mav23} propose to quantify cooperative strategies involving symmetry breaking using the \emph{winding number}~\cite{wnum}, a topological pairwise feature.
Specifically, \cite{Mav21} applies discrete winding numbers estimated between agents to distributed navigation, while \cite{Mav23} incorporates winding numbers into the cost function of robot controllers to maximize the absolute value of predicted winding numbers.
In this work, we choose to build our method upon the winding number because it provides a continuous topological representation that naturally accommodates scenarios with unobservable goals of other agents and can be seamlessly incorporated into a cost function of MPC.
% In this work, we choose to build upon the winding number for two key reasons. First, its ability to be treated as a continuous quantity makes it naturally applicable to scenarios where other agents' goals are unknown and facilitates its integration into an MPC cost function. Second, prior works~\cite{Mav21, Mav23} have already demonstrated its practical effectiveness. 
% In contrast, methods such as \cite{cao2019dynamic} are less suitable for the dense crossing scenarios we consider, as the graph's topology can change rapidly and unpredictably.

% Despite these contributions, challenges remain in how topological cooperative strategies are handled in existing winding number-based methods. First, both \cite{Mav21, Mav23} rely on hand-crafted cost functions or estimation models to select winding numbers, limiting flexibility in complex situations beyond what the designer anticipated. Moreover, treating winding numbers as discrete values reduces the expressiveness of cooperative strategies. Finally, \cite{Mav21} suffers from computational complexity $O(2^{n(n-1)/2})$ with respect to the number $n$ of agents, which hampers scalability. To tackle these problems, we introduce an approach that learns the strategies themselves---represented as continuous target winding numbers---through reinforcement learning, enabling flexible decision-making even where rule-based methods fail.

While winding number-based approaches are promising and have demonstrated practical effectiveness~\cite{Mav21, Mav23}, challenges remain in how topological cooperative strategies are handled. 
Specifically, \cite{Mav23} incorporates winding numbers into the cost function to maximize their absolute values. 
However, this sign-invariant objective renders mirrored topological choices equally preferable, leading to unstable switching in symmetric encounters. Furthermore, simply encouraging a large absolute winding number can induce unnecessary detours. 
On the other hand, \cite{Mav21} applies discrete winding numbers to represent topological relations. This approach suffers from computational complexity $O(2^{n(n-1)/2})$ with respect to the number of agents $n$, which hampers scalability. 
Additionally, these existing methods~\cite{Mav21, Mav23} rely on hand-crafted rules or discrete values, limiting the flexibility and expressiveness required for complex coordination (see Section~\ref{sec:results} for more discussion). 
To tackle these problems, we introduce an approach that learns the strategies themselves as continuous target winding numbers weighted by dynamic importance, enabling flexible decision-making.

Finally, we clarify the scope of our work. The winding number can also be used to represent collision-avoidance strategies with static obstacles~\cite{kuderer2014online}. While we focus on multi-agent navigation in an obstacle-free environment in our experiments, our framework is applicable to collision avoidance strategies with static obstacles.
% Furthermore, whereas several previous works using topological notions have considered human-robot interaction~\cite{Mav23,mavrogiannis2019multi,mavrogiannis2022implicit,cao2019dynamic,martinez2024shine}, 
% this paper concentrates on navigation among multiple robots that share a common policy.

\section{Preliminaries} \label{sec:preliminaries}
% We formulate the distributed multi-agent navigation problem and define winding number~\cite{wnum}, which describes the relation between agent trajectories.
% In this section, we formulate the navigation problem for a decentralized multi-agent system, and define the winding number\cite{wnum} to quantify the topological cooperative strategy governing how agents pass one another.

\subsection{Problem Formulation}
% Following~\cite{CADRL, SA-CADRL, GA3C}, distributed multi-agent navigation is modeled as a partially observable decision-making problem. 
Following~\cite{CADRL, SA-CADRL, GA3C}, distributed multi-agent navigation is modeled as a partially observable decision-making problem. In this paper, we use the term distributed navigation in the same sense as \cite{CADRL}, i.e., \emph{each agent knows nothing about other agents except their current positions, velocities and shapes}. Specifically, at each time step, each agent computes its own control by locally evaluating a policy; no centralized module solves a joint optimization over all agents or outputs actions for multiple agents.
Thus, in this work, distributed navigation refers to decentralized execution without centralized joint optimization or communication of goals, intentions, or planned trajectories, while assuming that the current observable states of other agents are available through sensing or state sharing.

Let $N \in \mathbb{Z}_{+}$ be the number of agents and $\mathcal{A}:=\{1,\ldots,N\}$ be the set of agent indices. The state of agent $i\in\mathcal{A}$ at time $k\in\mathbb{Z}_{\ge 0}$ is denoted by $\bs_k^i := [\bs_k^{i,o\top}, \bs_k^{i,h\top}]^\top$, where $\bs^o := [p_x, p_y, v_x, v_y, r] \in \mathbb{R}^5$ denotes observable quantities (position $\bp=[p_x,p_y]^\top$, velocity $\bv=[v_x,v_y]^\top$, radius $r$) and $\bs^h := [p_{gx}, p_{gy}, \theta]^\top \in \mathbb{R}^3$ denotes hidden quantities (goal position $\bp_{\mathrm{goal}}=[p_{gx},p_{gy}]$ and heading $\theta$). Let $\mathcal{U}\subset\mathbb{R}^2$ be the action space and $\bu_k^i\in \mathcal{U}$ the control input of agent $i$ at time $k$. All agents share the same deterministic dynamics $f:\mathbb{R}^9\times\mathbb{R}^2\to\mathbb{R}^9$:
\begin{equation}
    \bs_{k+1}^i = f(\bs_k^i, \bu_k^i), \ \ \ (\bs_{0}^i: \mathrm{given}, \ \forall i \in \cA).
\end{equation}
Each agent uses a shared policy $\pi$ such that $\bu_k^i \sim \pi\left( \bs_k^i, \by_k^i \right)$, where $\by_k^i := \{\bs_k^{j,o}\ |\ j \in \cA \setminus \{i\}\}$ denotes the observable states of others.
% The optimal control problem is:
The optimal control problem for a decentralized multi-agent system is formulated as follows.
\begin{prob}[Multi-Agent Navigation Problem]
Given initial states $\set{\bs_{i,0}}_{i \in \cA}$, find a memoryless policy $\pi$ that minimizes the expected time $k_g\in\mathbb{N}$ for all agents to reach their goals:
    \begin{align}
        \argmin_{\pi} \ & \mathbb{E}_\pi \left[ k_g\ |\ \set{\bs_{i,0}}_{i \in \cA}  \right]   \label{prob1-begin} \\ 
        \mathrm{s.t.}\ \ & \|\bp^i_k - \bp^j_k \|_2 > r_{i} + r_{j} \ \  (\forall i,j \in \cA, i\neq j), \label{prob1-2}\\
        \ \ & \bp^i_{k_g} = \bp^i_g \ \ \ \ (\forall i \in \cA),  \label{prob1-3}\\
        \ \ & \bs^i_{k+1} = f(\bs_k^i, \bu_k^i),\ \ \bu_k^i \sim \pi(\bs_k^i, \by_k^i) \ \ (\forall i \in \cA). \label{prob1-end}
    \end{align}
    \hfill$\triangle$ \label{prob:nav}
\end{prob}
% Solving Problem.\ref{prob:nav} exactly is intractable. As in many prior works (e.g., \cite{CADRL}), we assume a shared policy and learn it. 
% The key difficulty then lies in partial observability of others’ hidden states.
% The objective of this problem is to minimize the time required for all agents to reach their respective goals.
The objective of this problem is to minimize the time required for all agents to reach their respective goals. 
% This formulation is widely adopted in decentralized multi-agent navigation~\cite{CADRL, SA-CADRL, GA3C}. 
% The core difficulty of this setting stems from partial observability, which inherently arises from the inability to observe the goal positions of other agents or to obtain their accurate predicted trajectories.
The core difficulty of this setting stems from partial observability: each agent observes only the current observable states of other agents, while their goals, intentions, and future trajectories remain unavailable. The future trajectories used later in MPC are not privileged information, but are internally approximated from current states by constant-velocity extrapolation over a short receding horizon.

Note that, while this formulation assumes time synchronization among agents at each step $k$, our proposed framework does not strictly require it.

\subsection{Winding Number}
%Symmetry-induced deadlocks frequently occur in distributed navigation when agents occupy symmetric configurations. Breaking such symmetry requires behavior driven by a \emph{topological} passing strategy.
We adopt the winding number~\cite{wnum} as a quantitative feature for a \emph{topological} passing strategy. Following~\cite{Mav21, Mav23}, for two trajectories on the plane we define:

\begin{definition}[Winding Number]
Let $\theta_k^{i,j}$ be the bearing of agent $j$ from agent $i$ at time $k$. For the pair of observable trajectories $(\bs^{i,o}_{k:l}, \bs^{j,o}_{k:l})$ from time $k$ to $l$, the winding number $w$ is
    \begin{equation}
        w \left( \bs^{i,o}_{k:l}, \bs^{j,o}_{k:l} \right) := \frac{1}{2\pi} \sum_{\bar k=k}^{l-1} \Delta \theta_{\bar k}^{i,j}, \label{wnum}
    \end{equation}
where $\Delta \theta_k^{i,j} \in (-\pi, \pi]$ is the signed angle between $\theta_{k}^{i,j}$ and $\theta_{k+1}^{i,j}$.
\hfill$\Box$ \label{def:wnum}
\end{definition}
The sign of $w$ encodes the passing side, and its magnitude reflects progress toward passing. Selecting an appropriate target $w$ thus corresponds to choosing a specific passing strategy.
%By acting to make the predicted winding number match the target, an agent achieves the planned cooperative behavior.
Although $w$ is pairwise, prior work has shown it remains informative in scenarios with multiple simultaneous crossings~\cite{Mav21, Mav23}. 
% In our method, the Planner learns target values $w^{i,j}$ to enable cooperative symmetry breaking.

%%%%%%%%%%%%%%%%%%%%%%%%%%%%%%%%%%%%%%%%%%%%%%%%%%%%%%%%%%%%%%%%%%%%%%%%%%%%%%%%%%%%
%%%%%%%%%%%%%%%%%%%%%%%%%%%%%%%%%%%%%%%%%%%%%%%%%%%%%%%%%%%%%%%%%%%%%%%%%%%%%%%%%%%%

\section{Proposed Method} \label{sec:method}
In this section, we propose WNumMPC, a hierarchical method composed of a Planner and a Controller. The learning-based Planner $\pi^P_\theta$ \emph{learns} a topological strategy by outputting continuous-valued signed target winding numbers together with importance weights. The model-based Controller $\pi^C$ then safely executes this strategy via MPC, generating collision-free and efficient motions.

\subsection{Winding Number Planner} \label{sec:planner}
The learning-based Planner~$\pi^{\mathrm P}_\theta$ maps observations to a topological cooperative plan that breaks symmetry. 
% Concretely, it outputs target winding numbers relative to other agents.
% In addition, it determines the dynamic cost weights to indicate which interactions to prioritize, which enables flexible navigation even in dense environments.
Concretely, it outputs a target winding number $w_k^{i,j} \in [-1, 1]$ for every other agent $j \neq i$, together with an interaction weight $\alpha_{w,k}^{i,j} \in [0,1]$.
Rather than explicitly pruning agent pairs (e.g., agents far behind or already passed), our formulation keeps all pairs and lets the Planner down-weight irrelevant interactions through $\alpha_{w,k}^{i,j} \approx 0$, which effectively removes their influence in the cost \eqref{f_cost_w}.

Formally, the Planner is defined as follows:
\begin{gather}
    \bw_k^i, \boldsymbol{\alpha}_{w,k}^{i} \sim \pi_{\theta}^{\mathrm{P}}(\bs_k^i, \by_k^i),  \\
    \bw_k^i := \left(w_k^{i,j}\right)_{j \neq i},\ \ \boldsymbol{\alpha}_{w,k}^i := \left(\alpha_{w,k}^{i,j}\right)_{j \neq i}.
\end{gather}
Here $w_{k}^{i,j}\in [-1,1]$ is the target winding number for agent $i$ and $j$ from time $k$ to the end of the prediction horizon, and $\alpha_{w,k}^{i,j} \in [0,1]$ is the weight used in the cost terms of \eqref{f_cost}–\eqref{f_cost_w}.
The network parameters of the Planner are denoted by $\theta$.
Inputs $(\bs_k^i,\by_k^i)$ are represented in a rotation-invariant local frame, following~\cite{CADRL, SA-CADRL, GA3C}, to avoid geometric redundancy.

% We train $\pi^{\mathrm P}_\theta$ with PPO~\cite{PPO}. 
We train the planner policy $\pi^P_\theta$ with PPO~\cite{PPO} using a centralized critic under the centralized-training/decentralized-execution paradigm~\cite{CTDE}. Specifically, when updating agent $i$, the critic is conditioned on a joint state that includes agent $i$'s state $\bs^i_k$ and other agents' observable and hidden components $\{(\bs^{j,o}_k, \bs^{j,h}_k)\}_{j\neq i}$, which are represented in the same rotation-invariant local frame of agent $i$, while the planner is executed based only on $(\bs^i_k, \by^i_k)$.
Furthermore, all agents share parameters $\theta$, collect rollouts into a common rollout buffer, and optimize the shared planner. 
However, in this paper we train $ \pi^P_\theta$ separately for each agent count $N$.
Planning with only nearby agents is possible, but extending the Planner to variable-size neighborhoods and variable $N$ is left for future work.

During this training process, the reinforcement learning objective is given by
\begin{align}
    J(\pi) := \mathbb{E}_{\pi} \left[ \sum_{k=0}^{K_{\max}} \gamma^k r (\bs_k, \bu_k) \right],
\end{align}
with reward shaped as in~\cite{CADRL,SA-CADRL,GA3C}:
\begin{equation}
    r\left(\bs_k, \bu_k \right) 
    := \begin{cases}
        -1 & (d_{\min }<0) \\ 
        (d_{\min} - 0.25) / 2 & (d_{\min }<0.25) \\ 
        1 & (\mathbf{p} =\mathbf{p}_g) \\ 
        0 & (\text{otherwise}),
    \end{cases}
\end{equation}
where $d_{\min}$ is the distance to the nearest other agent, $\mathbf{p}$ the current position, and $\mathbf{p}_g$ the goal.
The discount factor is $\gamma$.
While the reward is calculated independently for each agent, episodes terminate for all agents upon any collision to encourage cooperative avoidance.
As in \cite{CADRL, SA-CADRL, GA3C}, we decay rewards quickly by setting $\gamma=0.95<1$ to encourage efficient navigation.

\subsection{Winding Number-aware Controller} \label{sec:controller}
The model-based Controller $\pi^{\mathrm C}$ produces actual control inputs $\bu_k^i$ that realize the planned topology, given the target winding numbers $\bw_k^i$ and weights $\boldsymbol{\alpha}_{w,k}^i$ of winding number costs.
Formally, $\bu_k^i = \pi^{\mathrm{C}} (\bs_k^i, \by_k^i ; \bw_k^i, \boldsymbol{\alpha}_{w,k}^i)$.
%It is implemented as MPC~\cite{MPC}.
% We implemented the Controller based on the MPC in \cite{poddar2023crowd}.
We implemented the controller using the MPC implementation from \cite{Mav23}, which was included in the source code of \cite{poddar2023crowd}.
At each step, we solve
\begin{align}
    \argmin_{\bar \bu_{0:K}\in\mathcal{U}}& \ \ \cJ\ (\bar \bs^i_{0:K}, \bar \by^i_{0:K} ; \bw^i_k, \boldsymbol{\alpha}_{w,k}^i) \label{prob:MPC} \\
    \mathrm{s.t.} & \ \ \bar \bs^i_{\bar k+1} = f(\bar \bs^i_{\bar k}, \bar \bu^i_{\bar k}) ,\ \ \bar \bs^i_0 = \bs^i_k,\ \ \bar \by^i_0 = \by^i_k,
\end{align}
and apply the initial action $\bu^{*,i}_0$ of the optimal sequence $\bu^{*,i}_{0:K}$ as $\bu_k^i$.
Here, $\mathcal{U}$ is the set of controls to consider, $K \in \dN$ is the prediction horizon, and future trajectories~$\bar{\by}_{\bar k}^i$ of other agents are approximated by constant-velocity extrapolation. 
This approximation is pragmatic in our setting because the horizon~$K$ is short 
and we replan in a receding-horizon manner, which limits error accumulation, so the induced prediction error is not severe in practice, as also reported in related navigation settings (e.g., \cite{Mav23, poddar2023crowd}). 
Importantly, the proposed framework does not rely on a specific prediction model: any trajectory prediction module can be plugged in as long as it provides $\overline{y}_{0:K}^{i}$. 
% In practice, residual prediction errors are mitigated by the receding-horizon replanning of MPC and by the learned topological strategy ($w_{k}^{i,j}$, $\alpha_{w,k}^{i,j}$) that provides robust symmetry-breaking guidance.

The cost of the Controller $\pi^{\mathrm C}$ is
\begin{align}
    & \cJ (\bar \bs^i_{0:K}, \bar \by^i_{0:K} ; \bw^i_k, \boldsymbol{\alpha}_{w,k}^i) := \alpha_g \cJ_g (\bar \bs^i_{0:K}) \\
    & \ + \alpha_o  \cJ_o (\bar \bs^i_{0:K}, \bar \by^i_{0:K}) + \cJ_w (\bar \bs^i_{0:K}, \bar \by^i_{0:K} ; \bw^i_k, \boldsymbol{\alpha}_{w,k}^{i}), \label{f_cost}
\end{align}
where $\cJ_g$ and $\cJ_o$ are penalty terms for goal reaching and collision avoidance, respectively, which are defined as follows:
\begin{align}
    \cJ_g (\bar \bs^i_{0:K}) & := \sum_{\bar k = 0}^{K} (\bar \bp^i_{\bar k} - \bar \bp^i_g)^\top Q_g (\bar \bp^i_{\bar k} - \bar \bp^i_g), \\
    \cJ_o (\bar \bs^i_{0:K}, \by^i_{0:K}) & := \sum_{j \in \cA \setminus \{i\}} \sum_{\bar k = 0}^{K} A_d^2 (\bar\bs^i_{\bar k}, \bar \bs^{j,o}_{\bar k}).
\end{align}
Here $Q_g\succ 0$, and $A_d$ denotes the asymmetric Gaussian integral function~\cite{AD}, dependent on the agent’s heading. The topology term penalizes deviation from the target winding numbers:
\begin{align}
    \cJ_w & \left(\bar \bs^i_{0:K}, \bar \by^i_{0:K} ; \bw^i_k, \boldsymbol{\alpha}_{w,k}^{i} \right) \label{f_cost_w} \\
    & :=  \frac{1}{N-1} \sum_{j \in \cA \setminus \{i\}} \alpha_{w,k}^{i,j} \left\{ w \left(\bar \bs^{i,o}_{0:K}, \bar \bs^{j,o}_{0:K} \right) - w_k^{i,j} \right\}^2. 
\end{align}
Since we assume policy homogeneity, we also use common values for the hyperparameters for $\pi^C$ among agents.
This design also provides scalability with respect to the number of agents $N$: each agent $i$ executes the controller independently by locally constructing $(s_k^i, y_k^i)$ and solving its own MPC problem~\eqref{prob:MPC}. 
Moreover, the interaction costs in \eqref{f_cost} are additive over other agents $j\in\mathcal{A}\setminus\{i\}$, and thus increasing $N$ increases the computation only linearly without any centralized joint optimization over all agents.

Our main novelty lies in the Planner; the Controller itself follows~\cite{Mav23}, modified only to incorporate the learned targets $\bw_k^i$ and weights $\boldsymbol{\alpha}_{w,k}^i$ via $\cJ_w$.
The specific design of the MPC algorithm, $\cJ_g$ and $\cJ_o$ are not the main part of this work and can be flexibly modified according to the implementation objectives.

\subsection{Overall Algorithm} \label{sec:wmpc}
The core idea is to delegate the symmetry-breaking topological decision to the learning-based Planner, while the model-based Controller ensures reliable local motion. 
% \cref{algo:wmpc} summarizes the procedure. 
% At each control step, every agent~$i \in \cA$ runs the Planner and the Controller locally using only 
% $(s_k^i,y_k^i)$, making the overall procedure decentralized and parallel across agents.
% % Here, $k$ indexes the MPC time steps; $\tilde{k}$ is an asynchronous counter controlling when agents refresh their winding-number plan.
% Here, $k$ indexes the control time steps; $\tilde{k}$ is an asynchronous counter that controls when each agent refreshes its high-level plan $(w_k^i, \alpha_{w,k}^i)$. 
% By randomly initializing $\tilde{k}$ for each agent, we de-synchronize their plan update timings to explicitly ensure that the proposed method operates asynchronously across the multi-agent system.

At each control step~$k$, every agent~$i \in \cA$ runs the Controller and outputs $\pi^{\mathrm{C}}(\bs_k^i, \by_k^i; \bw_k^i, \boldsymbol{\alpha}_{w,k}^i)$ as $\bu_k^i$, where $(\bs_k^i, \by_k^i)$ is the observation for agent~$i$ and $(\bw_k^i, \boldsymbol{\alpha}_{w,k}^i)$ is the output of the Planner. Agent~$i$ also runs the Planner and updates the value of $\bw_k^i,\boldsymbol{\alpha}_{w,k}^i$ by $\pi^{\mathrm{P}}_\theta (\bs_k^i, \by_k^i)$ once per $\tilde K$ time steps. In steps where no update occurs, $\bw_{k-1}^i,\boldsymbol{\alpha}_{w,k-1}^i$ are assigned to $\bw_k^i,\boldsymbol{\alpha}_{w,k}^i$.
The update timing for agents is desynchronized as follows. First, a random value $\tilde k_i\in\{0,1,\ldots,\tilde K-1\}$ is selected for each agent~$i$, and the Planner runs when $k=0$ or $k\equiv \tilde{k}_i \ (\bmod \tilde{K})$.

% \begin{figure}[!t]
%     \begin{algorithm}[H]
% 	   \begin{algorithmic}[1]
%         \Require{$\tilde K$, $\pi^{\mathrm{P}}_\theta$, $\pi^{\mathrm{C}}$}
%         \Ensure{trajectory of Agent~$i$ $\bs_{0:k_f}^i$}
%         \State{$k\leftarrow 0$}
%         \State{$\tilde{k} \leftarrow \mathrm{Random}(0,...,\tilde{K}-1)$}
%         \While{not reached goal}
%         \State{$\bs_k^i, \by_k^i \leftarrow \mathrm{getObservation}$}
%         \If{$k = 0$ or $\tilde k \equiv 0 \ (\bmod \ \tilde K)$}
%         \State{$\bw_k^i \leftarrow \pi^{\mathrm{P}}_\theta (\bs_k^i, \by_k^i)$}
%         \Else{}
%         \State{$\bw_k^i \leftarrow \bw_{k-1}^i$}
%         \EndIf
%         \State{$\bu_k^i \leftarrow \pi^{\mathrm{C}}(\bs_k^i, \by_k^i, \bw_k^i)$}
%         \State{$ k\leftarrow k+1$, $\tilde k \leftarrow \tilde k + 1$}
%         \EndWhile
% 	   \end{algorithmic} 
% 	   \caption{WNumMPC (Winding Number-aware MPC)}
%       \label{algo:wmpc}
%     \end{algorithm}
% \end{figure}

%%%%%%%%%%%%%%%%%%%%%%%%%%%%%%%%%%%%%%%%%%%%%%%%%%%%%%%%%%%%%%%%%%%%%%%%%%%%%%%%
%%%%%%%%%%%%%%%%%%%%%%%%%%%%%%%%%%%%%%%%%%%%%%%%%%%%%%%%%%%%%%%%%%%%%%%%%%%%%%%%

\section{Experiments}
We evaluated WNumMPC (our method) in simulation against ORCA~\cite{ORCA}, CADRL~\cite{CADRL}, Vanilla MPC, and T-MPC~\cite{Mav23} baselines.
Vanilla MPC is the MPC method that does not use winding numbers (i.e., \eqref{f_cost} with $\boldsymbol{\alpha}_{w,k}^i = 0$).
T-MPC~\cite{Mav23} is the MPC method with cost encouraging higher absolute values of winding numbers, without the dynamic weighting
(i.e., \eqref{f_cost} with $\boldsymbol{w}_k^{i,j}=0$ and $\boldsymbol{\alpha}_{w,k}^{i,j} = \mathrm{const.}<0$).
We also conducted real-robot experiments comparing the proposed method with Vanilla MPC and T-MPC to validate effectiveness in the physical world.

\subsection{Simulation Experiments}
\subsubsection{Dynamics}
In simulation, agents are holonomic with maximum speed $0.8$. The input $\bu=(u_x,u_y)$ directly sets the $x$ and $y$ velocities with $\| \bu \|\le 0.8$:
\begin{align}
&f((p_x, p_y, v_x, v_y, \theta, r), (u_x, u_y)) \\
=& (p_x+u_x\Delta t, \, p_y+u_y\Delta t, \, u_x, \, u_y, \, \tan^{-1}(u_y/u_x), \, r),
\end{align}
with control period $\Delta t=0.1$. All agents have collision radius $r_i=0.15$.

\subsubsection{Instances and Metric}
We consider dense scenarios where multiple agents cross simultaneously. Start positions are randomly generated around a circle.
More precisely, they are generated by adding random noise to randomly generated positions on the circle.
Goals are generated in two ways:
\begin{itemize}
\item \textbf{Random}: goals are sampled in the same way as the starts.
\item \textbf{Crossing}: each goal is placed diametrically opposite the start across the circle center.
\end{itemize}
The circle radius is $2.0$ and perturbations are drawn uniformly from $[-0.4,0.4]$ on each axis. 
An example of a Crossing instance is shown in \cref{fig:trajectories}.
We evaluate $100$ episodes for each agent count $N$ and each instance type in holonomic simulation.

To evaluate methods, we used the \emph{Average Extra Time to Goal}:
\begin{align}
  \bar{t}_e = \frac{1}{N} \sum_{i=1}^N \left[ t^i_{g} - \frac{\| \bp^i_{0} - \bp^i_{g} \|_2}{v_{\max}} \right],
\end{align}
a standard metric independent of agent count and start–goal distances~\cite{CADRL}. Instances are regarded as a timeout if not all agents reach goals within $20$ time units.

\subsubsection{Implementation Details}
Our implementation is based on \cite{Mav23, poddar2023crowd}. We used Python~3.11 and PyTorch.
The Planner is trained with TorchRL~\cite{torchrl}, utilizing its TanhNormal distribution to constrain the policy outputs, 
which restricts the target winding numbers $w_{k}^{i,j}$ to $[-1, 1]$ and the interaction weights $\alpha_{w,k}^{i,j}$ to $[0, 1]$. 
Both the actor and the critic are MLPs with three hidden layers and $\tanh$ activations, 
and the actor uses $128$ units per layer and the critic uses $256$ units per layer.
% PPO hyperparameters: entropy coefficient $1.0\times 10^{-4}$, learning rate $3.0\times 10^{-4}$, discount factor $0.95$, GAE $\lambda=0.9$, clipping threshold $0.2$, batch size $128$, epochs $15$.
We used the following PPO hyperparameters: a discount factor of $0.95$, GAE $\lambda=0.9$, a clipping threshold of $0.1$, and $4$ epochs. 
The learning rate, batch size, and entropy coefficient were set to $2.0 \times 10^{-4}$, $256$, and $1.0 \times 10^{-2}$, respectively. 
% For scenarios with $N > 5$ agents, these values were adjusted to $2.0 \times 10^{-4}$, $4096$, and $3.0 \times 10^{-3}$, respectively.
Unlike~\cite{CADRL, SA-CADRL, GA3C}, we train $\pi^P_\theta$ from scratch without behavioral pretraining; the model-based Controller already provides basic navigational competence early in training.

% Controller settings: horizon $K=10$, update period $\bar K=5$, $A_d$ parameters $\sigma_h=0.5$, $\sigma_r=0.3$, $\sigma_s=0.35$. We tuned $\alpha_g,\alpha_o$ for each number of agents and for both Vanilla MPC and T-MPC, by using Optuna\cite{optuna} with randomly generated Random instances. For Wnum MPC, we used the same parameters as Vanilla MPC.

Controller~$\pi^C$ parameters were set as follows:
prediction horizon $K=10$, update period $\tilde K=5$ and the shape parameters for $A_d$, $\sigma_h=0.5$, $\sigma_r=0.3$, $\sigma_s=0.35$.
We set $\mathcal{U}$ as a set of controls corresponding to constant velocities selected from $5$ speeds and $16$ evenly divided angles.
For WNumMPC, T-MPC, and Vanilla MPC, the cost function weights $\alpha_g$ and $\alpha_o$ were tuned for each experimental setup, 
such as the number of agents, using Optuna \cite{optuna}.

For training of both WNumMPC and CADRL, we used only Random instances.
% Training was conducted for each number of agents.
We trained for each number of agents with $129{,}600$ total training episodes, counting each agent trajectory as one episode.
For CADRL, the model was first pretrained by imitation on ORCA trajectories for $N=3$, then tuned with RL for each number of agents.
(This recipe was more workable than doing imitation learning for more than three agents.)
All baselines are run under the same computational environment with reasonable parameter tuning for fairness.

\subsection{Real-World Experiments}
We conducted tabletop experiments with differential-drive robots \emph{maru}~\cite{ichihashi2024swarm}, comparing our method with Vanilla MPC and T-MPC. CADRL and ORCA were omitted due to frequent collisions or timeouts in simulation.
Maru is a miniature-sized robot equipped with an IMU and can calculate its own position by reading the projected coded-pattern.

\subsubsection{Dynamics}
Since maru is differential-drive, we adapt the dynamics of agents, which is used in both the Planner’s training and the MPC model. The input $\bu=(u_v,\psi)$ denotes linear and angular velocity:
\begin{align}
&f((p_x, p_y, v \cos \theta, v\sin \theta, \theta, r), (u_v, \psi)) \\
=& (p_x+u_v \cos (\theta+\psi\Delta t/2)\Delta t,\ p_y+u_v\sin(\theta+\psi\Delta t/2)\Delta t, \\
&\ \ u_v \cos(\theta+\psi\Delta t), u_v \sin(\theta+\psi\Delta t), \theta + \psi \Delta t, r).
\end{align}
The control interval is $\Delta t=0.1$ seconds.
Inputs satisfy $|u_v + \psi/7.5|\le 0.6$ and $|u_v - \psi/7.5|\le 0.6$~\cite{ichihashi2024swarm}.
Thus, the max linear speed is $0.6$ per second in simulator units.
The collision radius is $r_i=0.15$, which corresponds to the physical robot radius $15$ mm. We use this value for collision detection in the experiments.
On the other hand, to mitigate sim-to-real discrepancies, we use $r_i=0.20$ in both the training environment and MPC cost computation.

\subsubsection{Instances and Metric}
We set $N=7$ and evaluate $400$ episodes (200 Random + 200 Crossing), alternating them so that goals of one instance serve as starts of the next for continuous operation. The circle radius is $1.5$ (about 15 cm), with axis-wise uniform noise in $[-0.3,0.3]$. The metric is the same as in the holonomic simulation. The timeout threshold is $30$ seconds.

\subsubsection{Implementation Details}
% Training settings mirror simulation.
The Planner policy $\pi^{\mathrm{P}}_\theta$ was trained with PPO entirely in a simulation environment 
using a differential-drive robot model. 
This policy was subsequently deployed on the physical hardware for zero-shot evaluation,
without any fine-tuning or further training on real-world data. 
We set $\mathcal{U}$ as a set of controls that take a constant action where each of the left and right wheels takes one of nine possible signed angular velocities.
All other training configurations and hyperparameters were identical to those used in the simulation.

\begin{figure*}[t]
    \centering
    \includegraphics[width=0.90\linewidth]{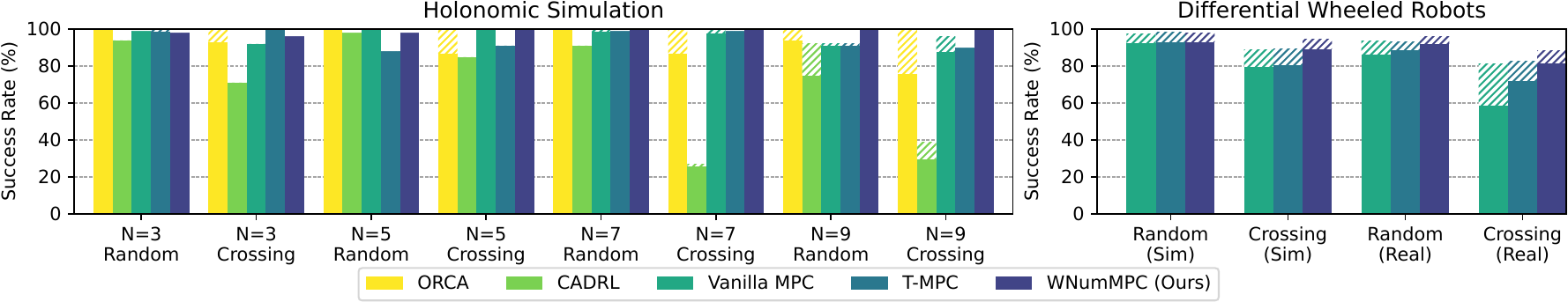}
    \caption{Comparison of Navigation Success Rates (solid bars) and Timeout Rates (hatched bars).
(Left) Results in simulations of the holonomic model for each agent count ($N$) and generation method of instances (Random, Crossing).
(Right) Results of the MPC-based methods with $N=7$ differential wheeled robots in simulations and real-world experiments.}
    \label{success_rate}
\end{figure*}

\begin{figure*}[t]
    \centering
    \includegraphics[width=0.90\linewidth]{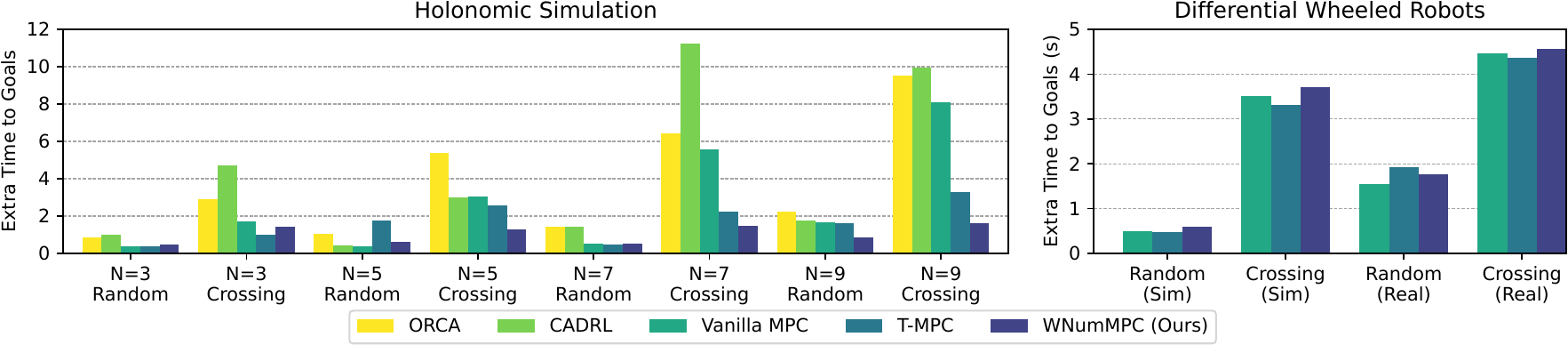}
    \caption{Comparison of Navigation Efficiency Based on average Extra Time to Goal.
(Left) Average extra time in holonomic model simulations for each agent count (N) and generation method of instances (Random, Crossing).
(Right) Average extra time for MPC-based methods with $N=7$ differential wheeled robots in simulations and real-world experiments.}
    \label{ave_etg}
\end{figure*}

\section{Results} \label{sec:results}
%Based on the evaluation metrics of the experimental results, we performed a quantitative analysis of the proposed method and compared its performance with that of existing methods.
We performed quantitative and qualitative comparisons of the experimental results.

\subsection{Simulation Results}
%\subsubsection{Performance Comparison with Baselines}

\cref{success_rate,ave_etg} (left) report success rate and average extra time to goal across agent counts $N$ and instance types. Existing methods degraded notably for larger values of $N$, especially on Crossing instances, whereas WNumMPC maintained high success rates even in dense settings. Moreover, while baselines incurred significantly larger extra time on Crossing, our method kept it low, indicating efficient navigation. These results indicate that WNumMPC effectively accounts for interactions and breaks symmetry in complex, dynamic environments.
% CADRL showed relatively good performance on Random instances except in the densest cases, but its performance degraded even in low-density Crossing instances. This is likely because training was conducted only on Random instances, leading to poor generalization in atypical scenarios. In contrast, our proposed method also trained solely on Random instances, yet still achieved favorable results on Crossing instances. This suggests that the reliability of the MPC component compensates for the shortcomings of purely learning-based approaches.

In contrast, as indicated by the success rates in \cref{success_rate}, CADRL resulted in a higher frequency of collisions compared to other methods.
While CADRL showed relatively good performance on Random instances with a small number of agents, its performance degraded even in low-density Crossing instances.
% Crossing instances tend to involve simultaneous interactions among multiple agents more frequently than Random instances, creating collision-prone scenarios where ensuring safety is challenging for methods relying solely on a learned policy.
% Conversely, our method achieved favorable results on Crossing instances, even though it was also trained solely on Random instances.
% This result suggests that the reliability of the MPC component effectively compensates for the safety shortcomings of purely learning-based approaches.
Crossing instances tend to involve simultaneous interactions among multiple agents more frequently than Random instances, creating collision-prone scenarios. Consequently, CADRL relying solely on a learned policy struggled to ensure safety, resulting in frequent collisions. In contrast, MPC-based methods maintained low collision rates even in such crossing scenarios, with the proposed method, in particular, achieving strong results.
These results suggest that the reliability of the MPC component effectively compensates for the safety shortcomings of purely learning-based approaches.

% \cref{fig:trajectories} visualizes trajectories for the same Crossing instance.
% In ORCA, agents stopped after yielding to each other. In CADRL, agents failed to avoid collisions after gathering in the center. Vanilla MPC successfully avoided collisions, but many agents temporarily stopped. T-MPC avoided collisions without slowing down, but some agents took extremely long detours. In contrast, WNumMPC successfully avoided collisions efficiently.

\begin{figure*}[t]
  \centering
  \subfloat[ORCA\cite{ORCA}]{%
    \includegraphics[width=0.17\textwidth]{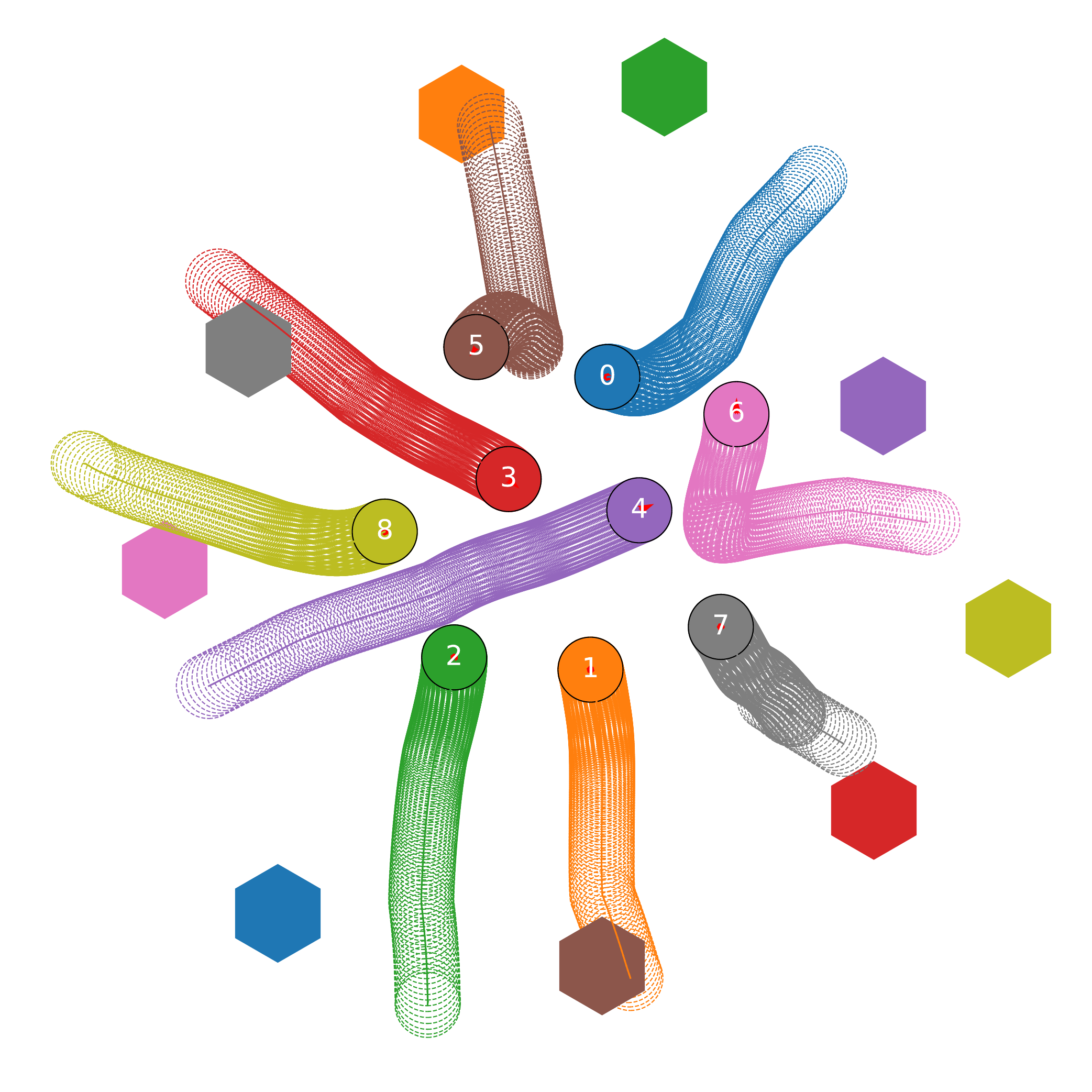}%
    \label{fig:orca}%
  }
  \hfill
  \subfloat[CADRL\cite{CADRL}]{%
    \includegraphics[width=0.17\textwidth]{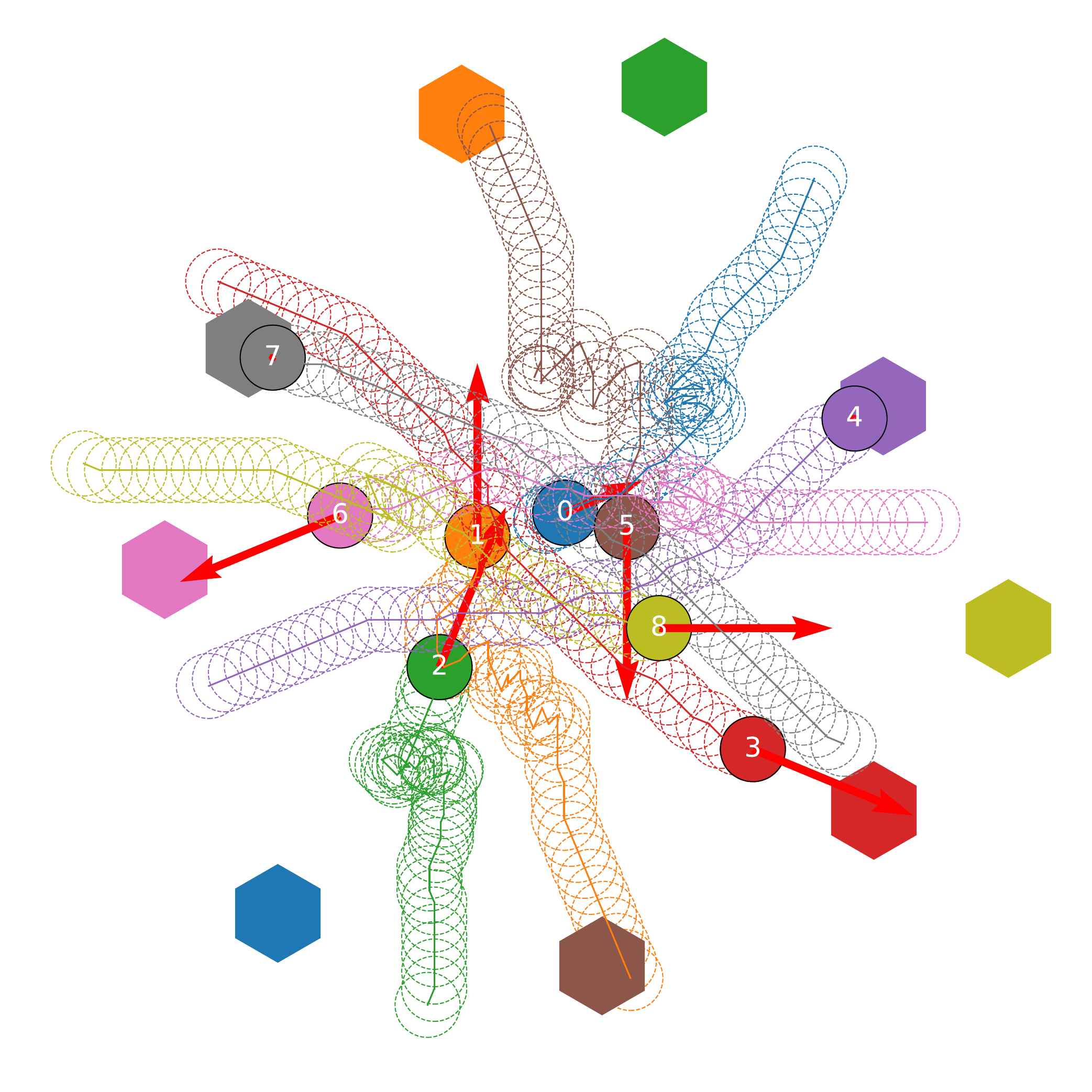}%
    \label{fig:cadrl}%
  }
  \hfill
  \subfloat[Vanilla MPC]{%
    \includegraphics[width=0.17\textwidth]{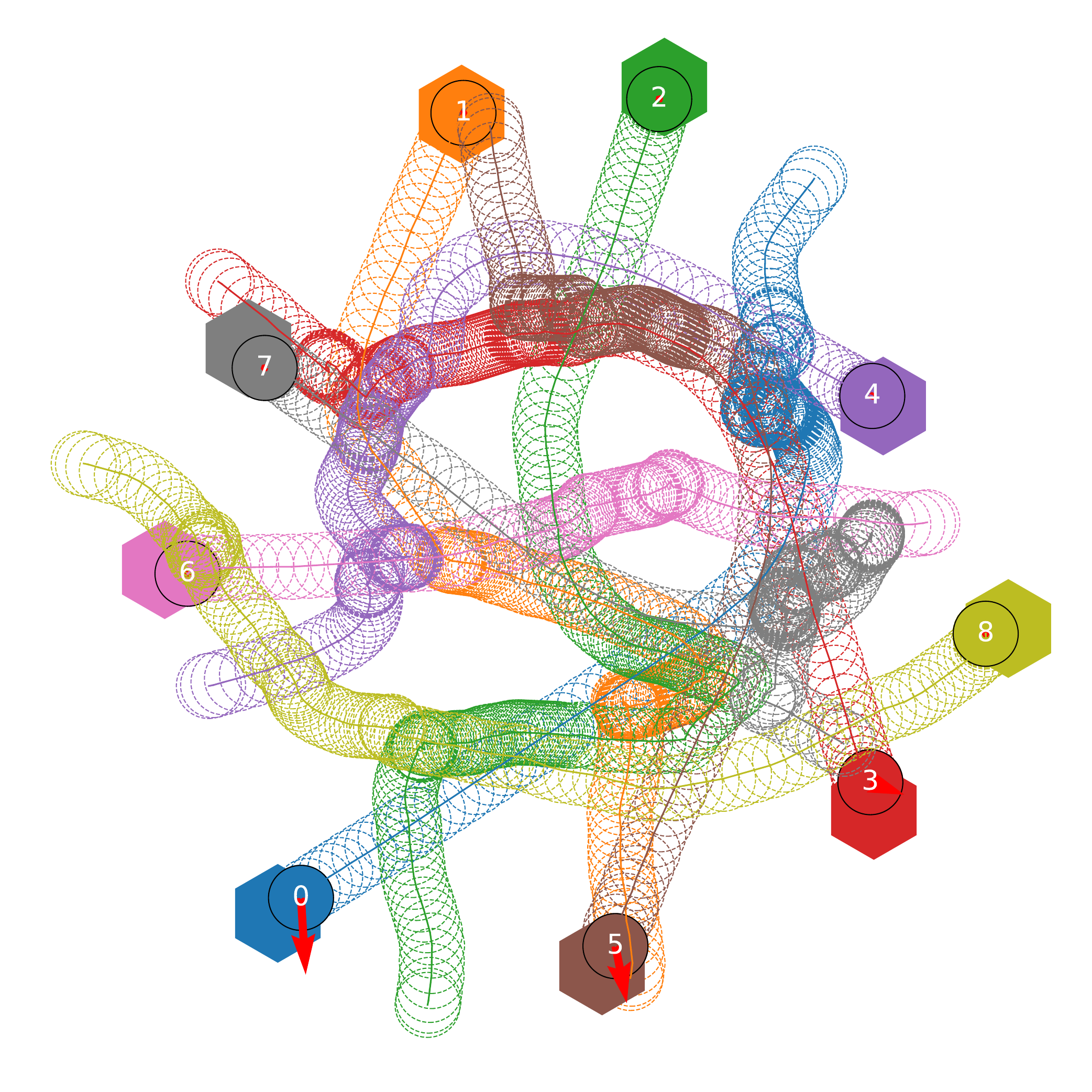}%
    \label{fig:normal_mpc}%
  }
  \hfill
  \subfloat[T-MPC\cite{Mav23}]{%
    \includegraphics[width=0.17\textwidth]{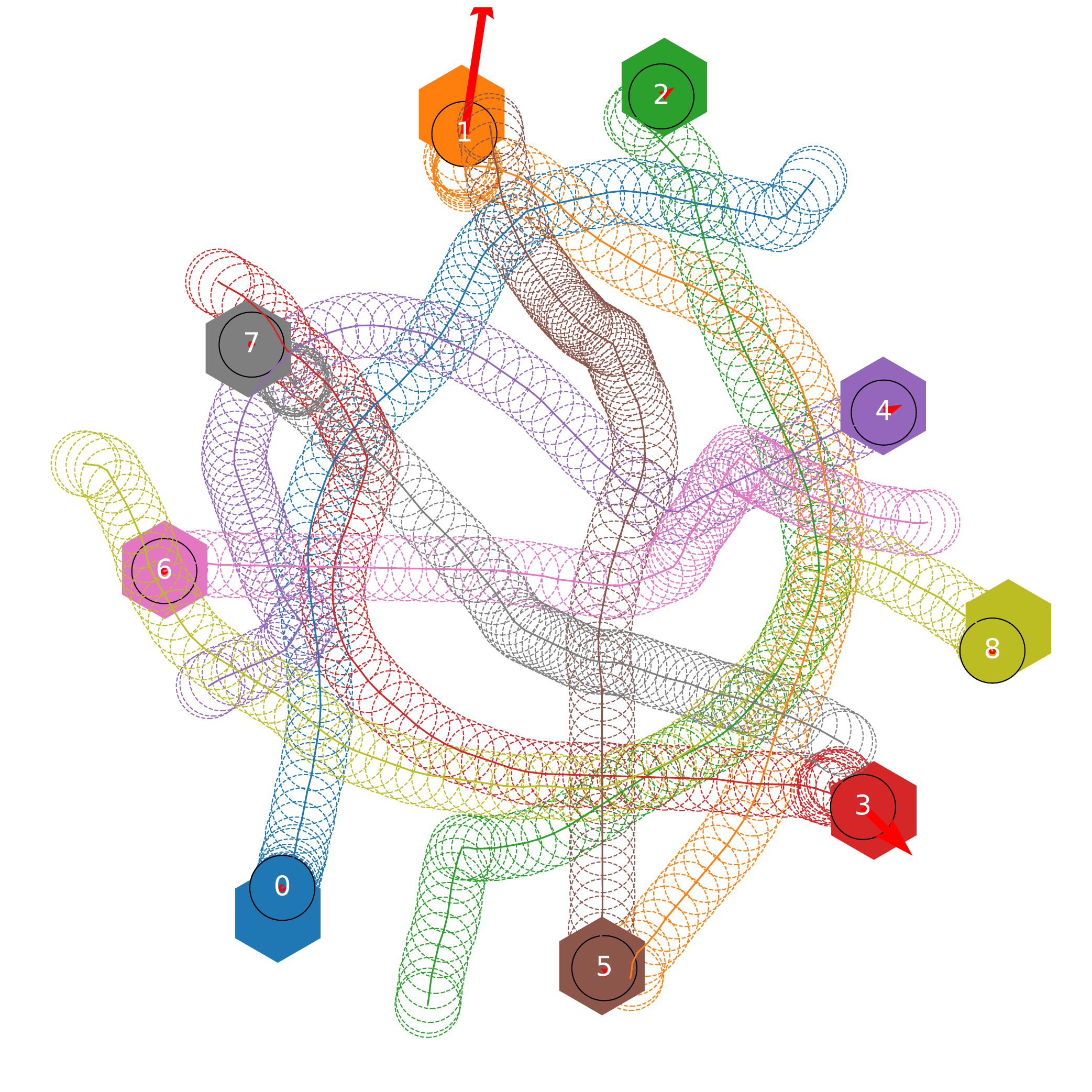}%
    \label{fig:mean_mpc}%
  }\hfill
  \subfloat[WNumMPC (Ours)]{%
    \includegraphics[width=0.17\textwidth]{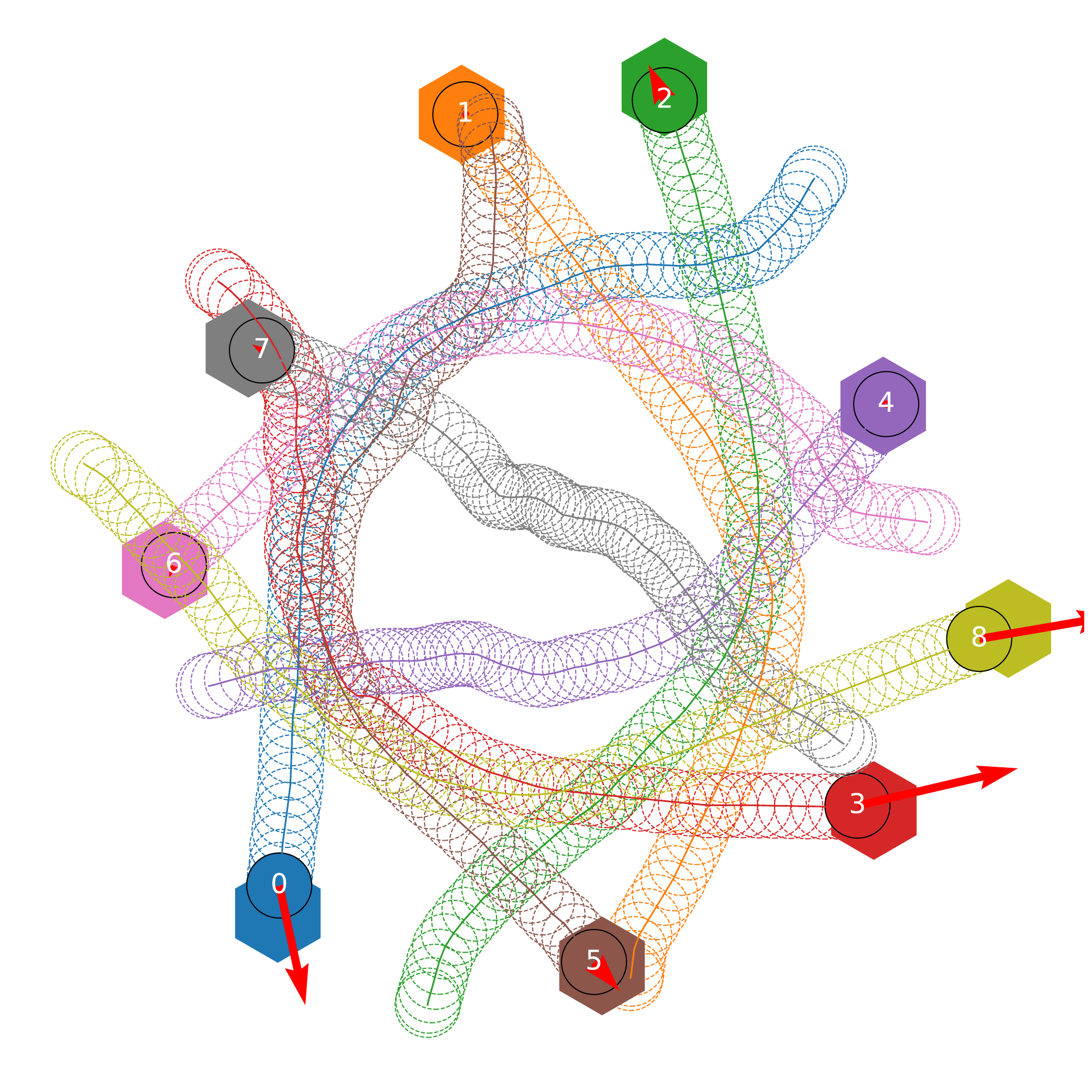}%
    \label{fig:wnum_mpc}%
  }
  \caption{
        % Comparison of agent trajectories in a Crossing scenario, 
        % generated by (a) ORCA, (b) CADRL, (c) Vanilla MPC, (d) T-MPC, and (e) WNumMPC.
        Trajectories in the same Crossing scenario with $9$ agents for (a) ORCA, (b) CADRL, (c) Vanilla MPC, (d) T-MPC, and (e) WNumMPC.
        The proposed method (e) efficiently breaks symmetry, 
        unlike baseline methods that result in deadlocks (a), collisions (b), and inefficient paths (c, d).
  }
  \label{fig:trajectories}
\end{figure*}

\begin{figure*}[t]
  \centering
  \subfloat[
        Temporal evolution of the T-MPC\cite{Mav23} trajectories in a scenario with $N=7$ agents.
        All agents successfully reach their goals at $t=9.9$.
  ]{%
    \includegraphics[width=0.15\textwidth]{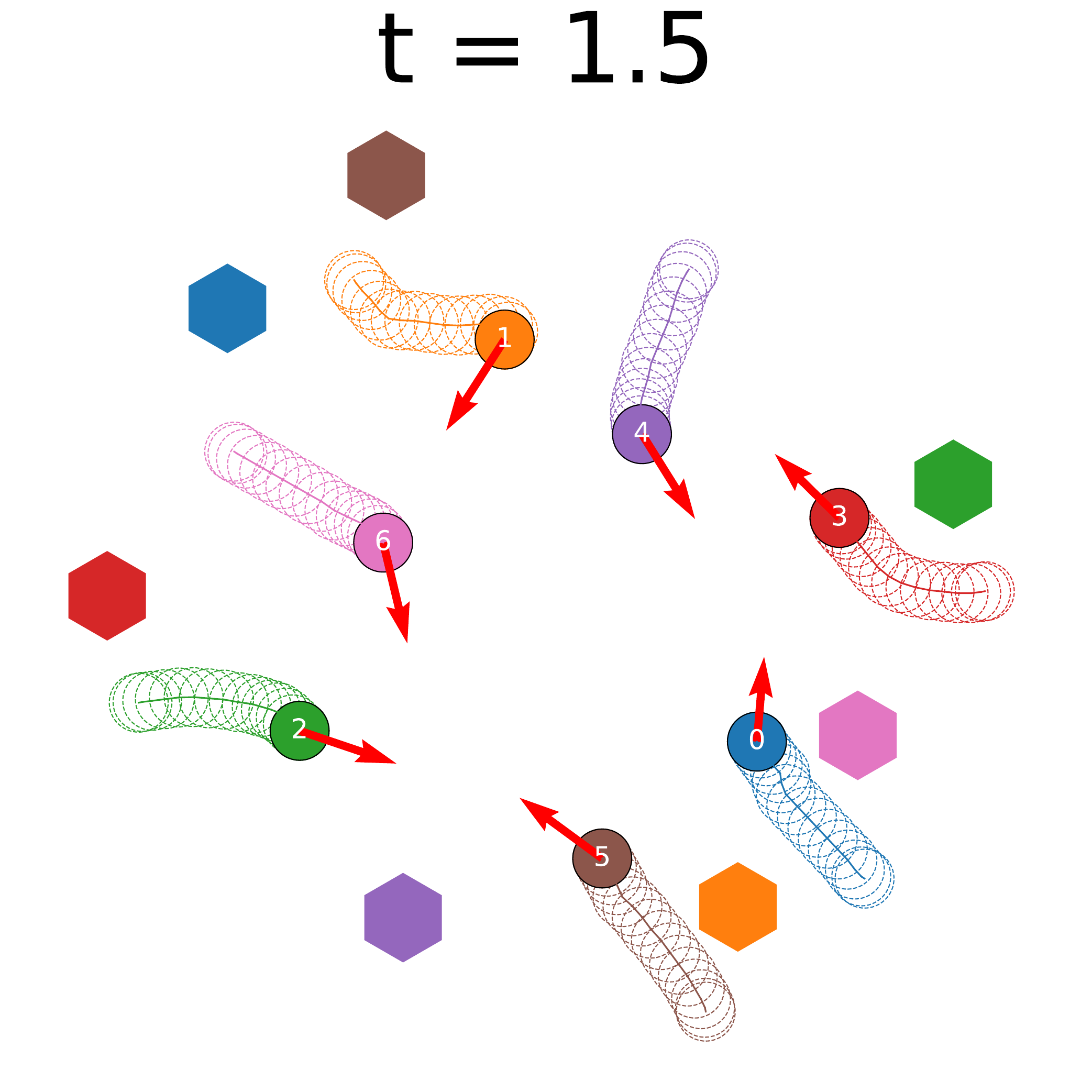}%
    \hfill
    \includegraphics[width=0.15\textwidth]{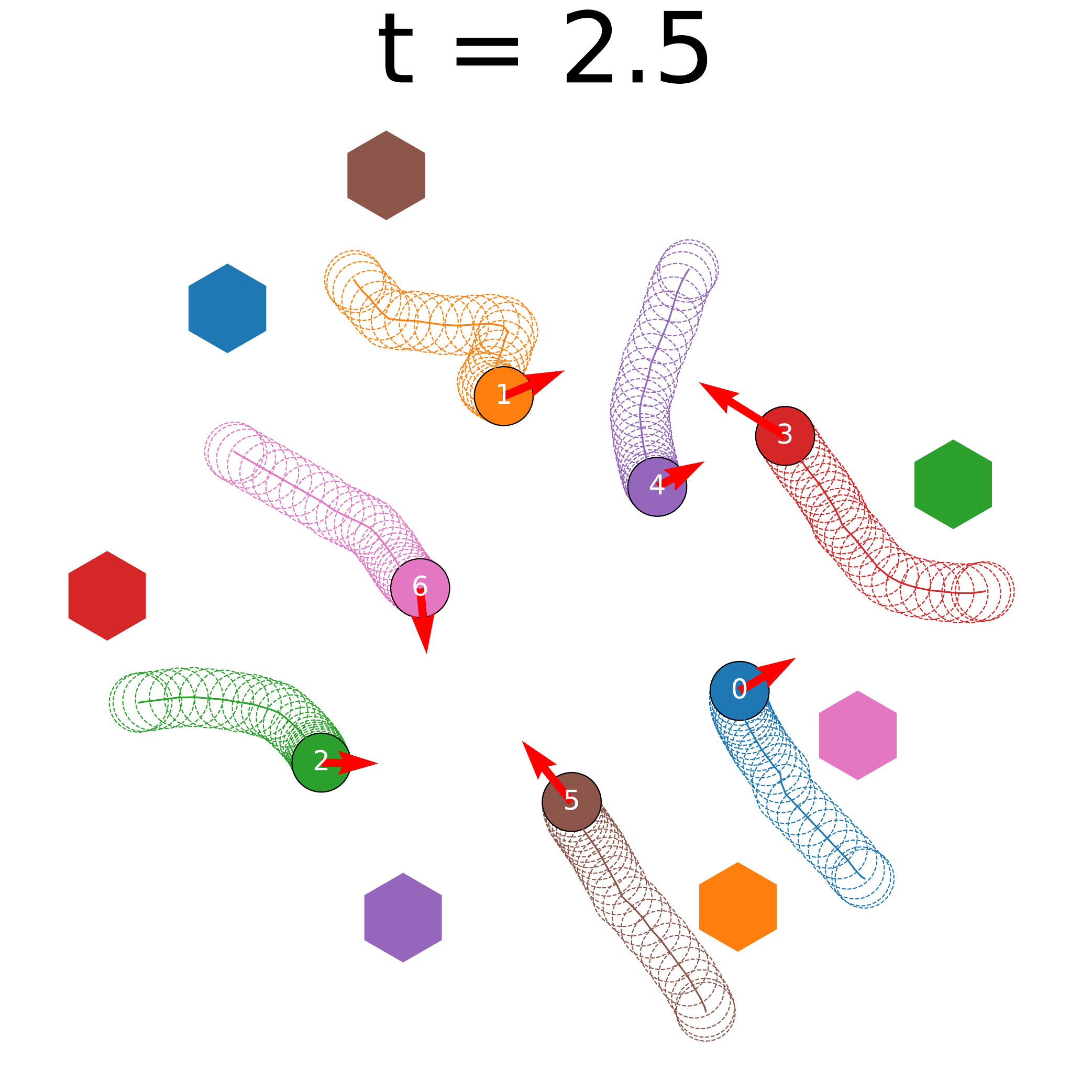}%
    \hfill
    \includegraphics[width=0.15\textwidth]{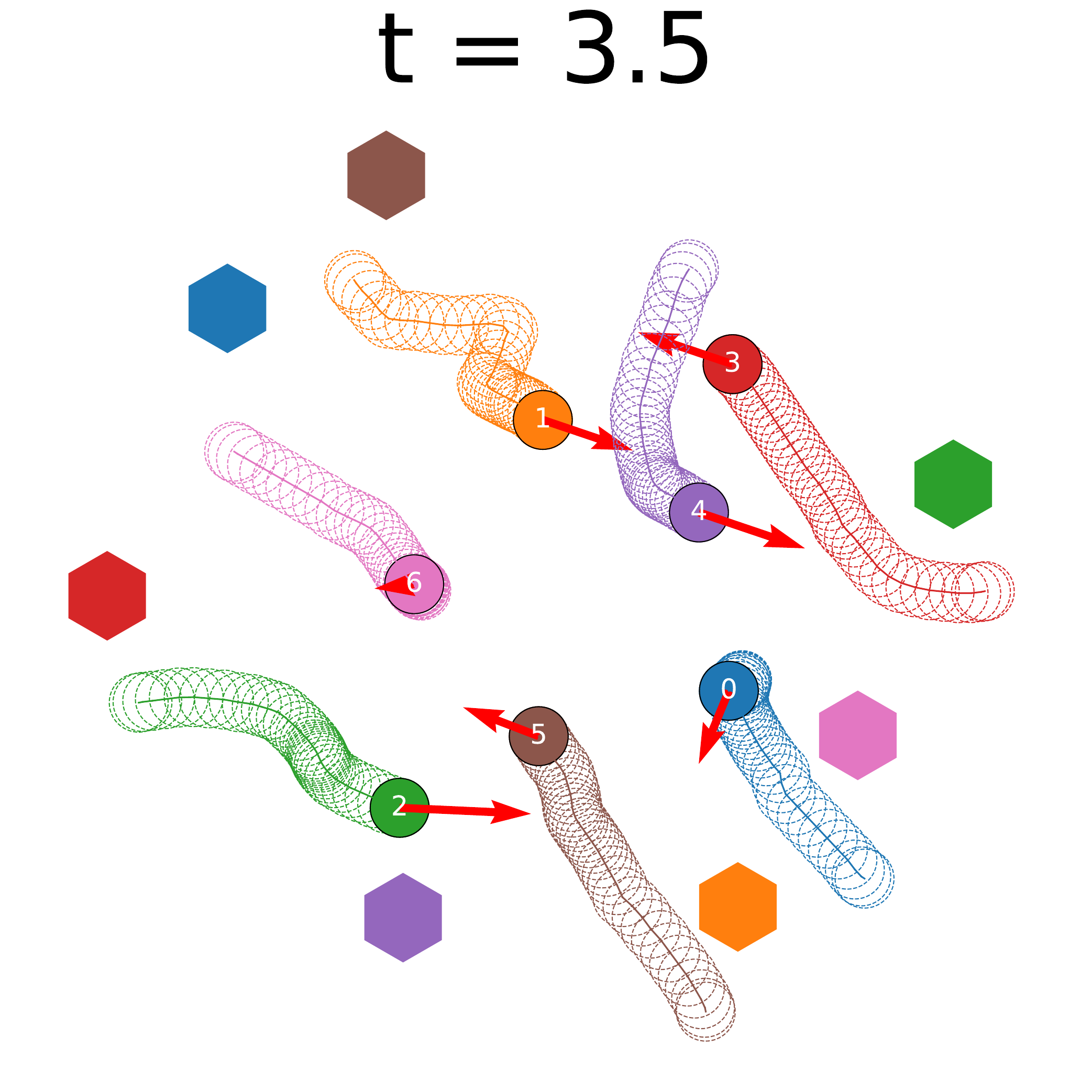}%
    \hfill
    \includegraphics[width=0.15\textwidth]{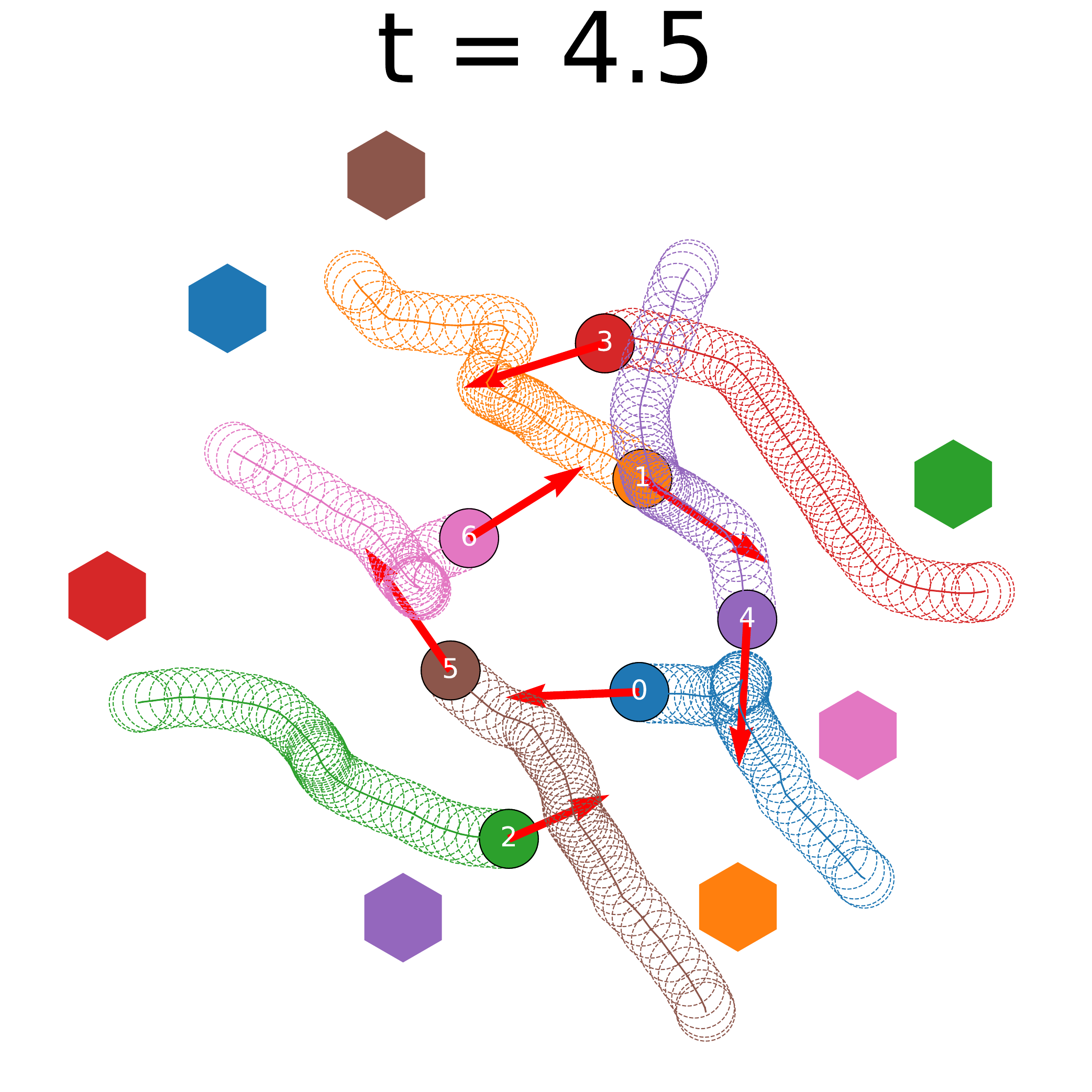}%
    \hfill
    \includegraphics[width=0.15\textwidth]{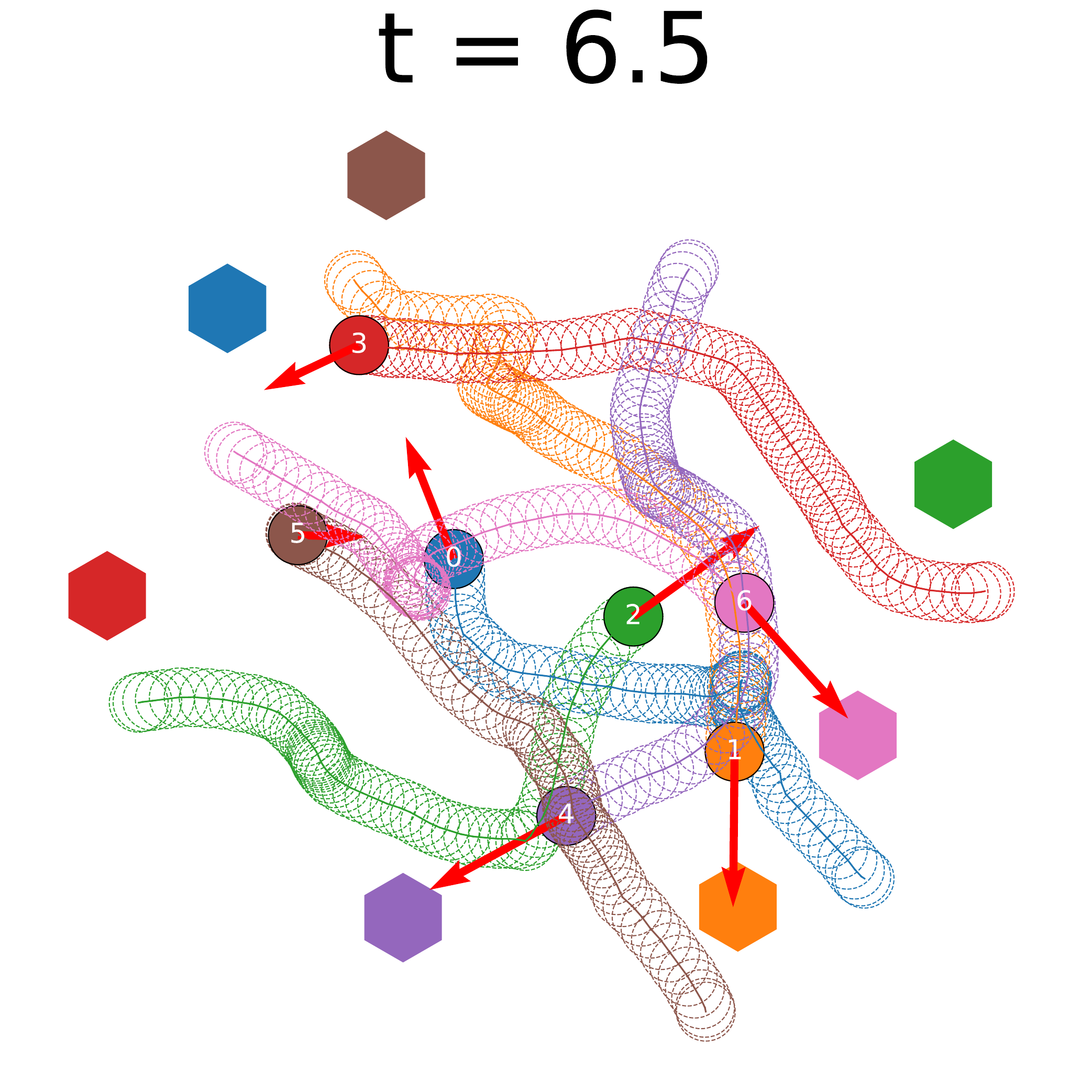}%
    \hfill
    \includegraphics[width=0.15\textwidth]{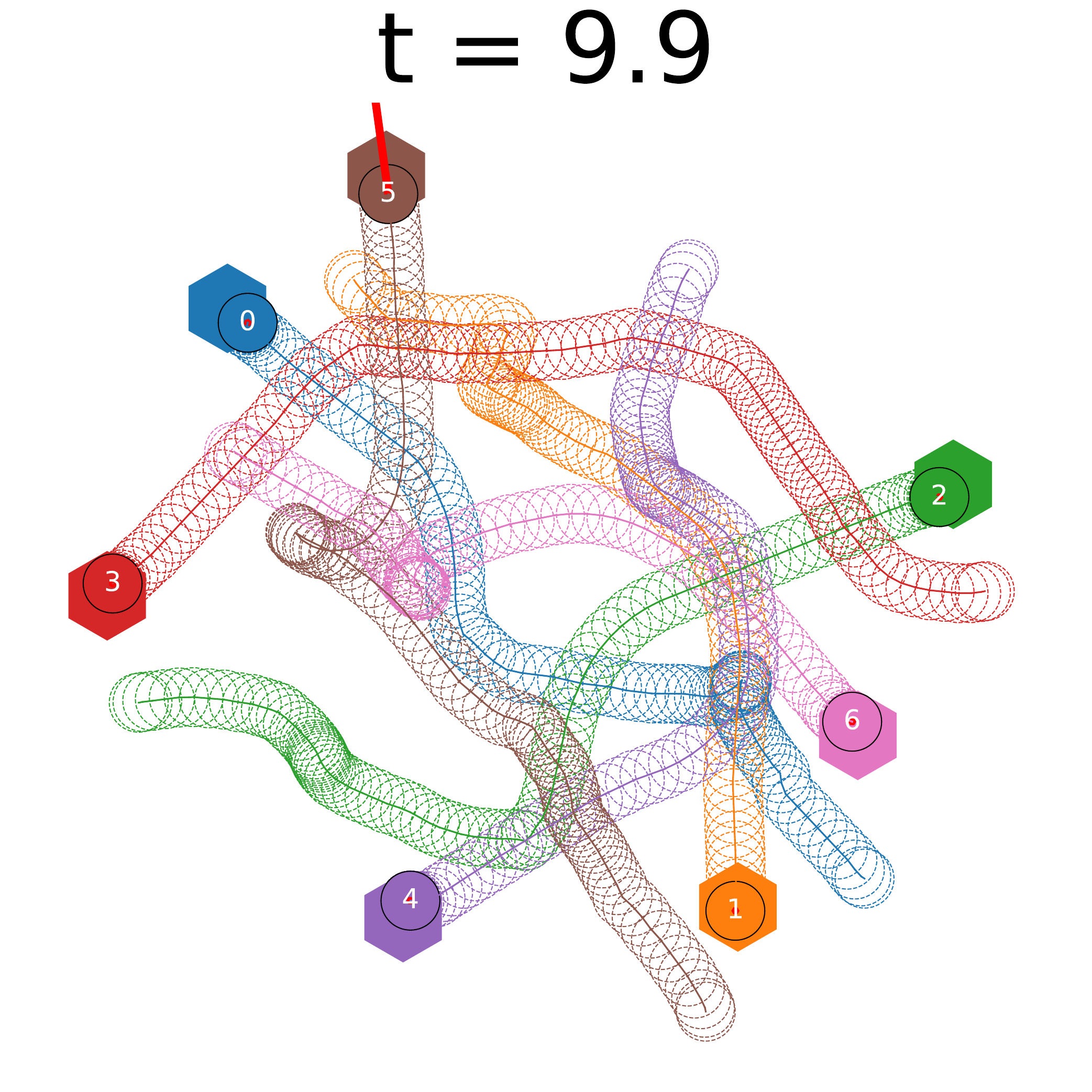}%
    \label{fig:traj_compare:tmpc}
  }
  \\
  \subfloat[
        Temporal evolution of the WNumMPC trajectories in a scenario with $N=7$ agents.
        All agents successfully reach their goals at $t=7.4$.
  ]{%
    \includegraphics[width=0.15\textwidth]{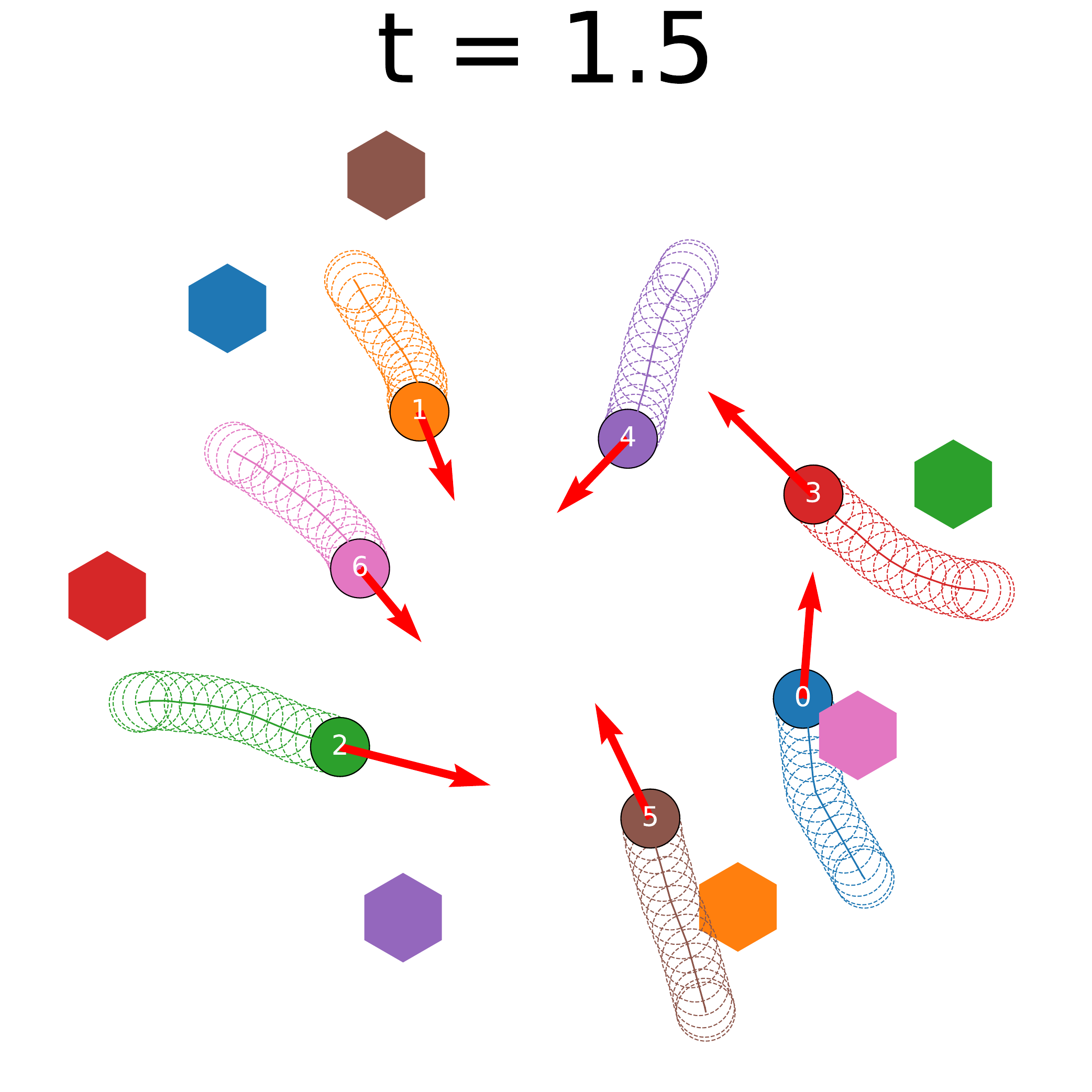}%
    \hfill
    \includegraphics[width=0.15\textwidth]{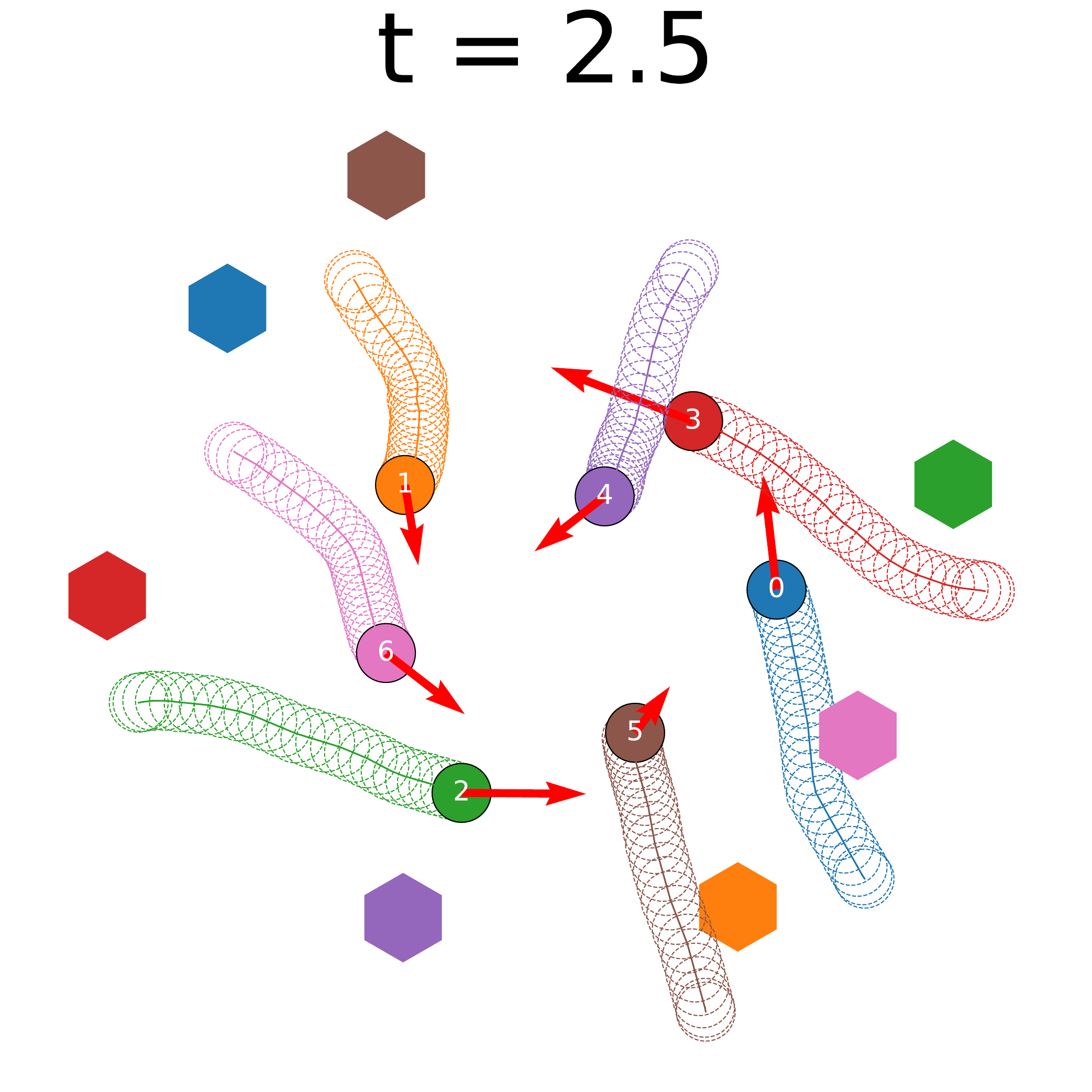}%
    \hfill
    \includegraphics[width=0.15\textwidth]{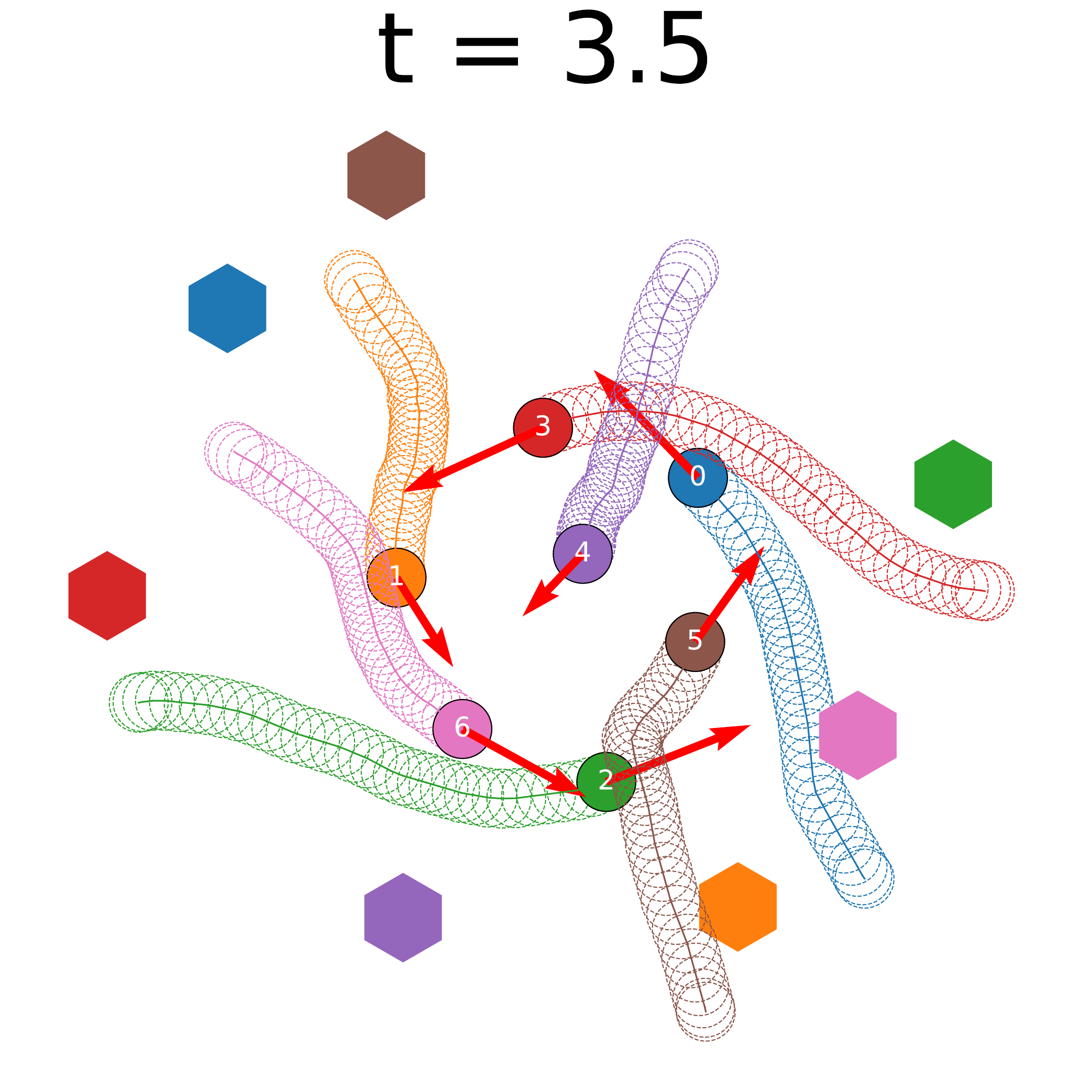}%
    \hfill
    \includegraphics[width=0.15\textwidth]{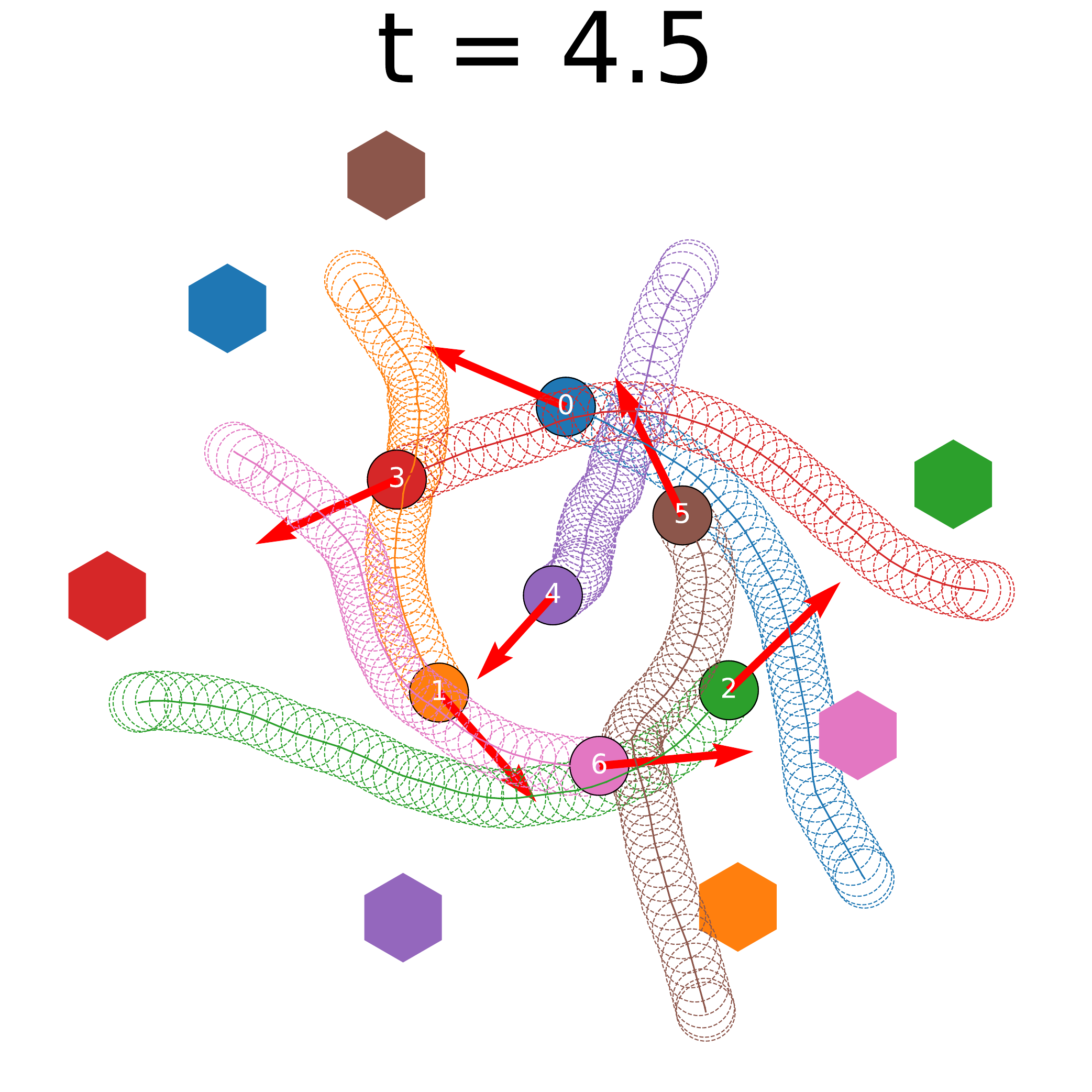}%
    \hfill
    \includegraphics[width=0.15\textwidth]{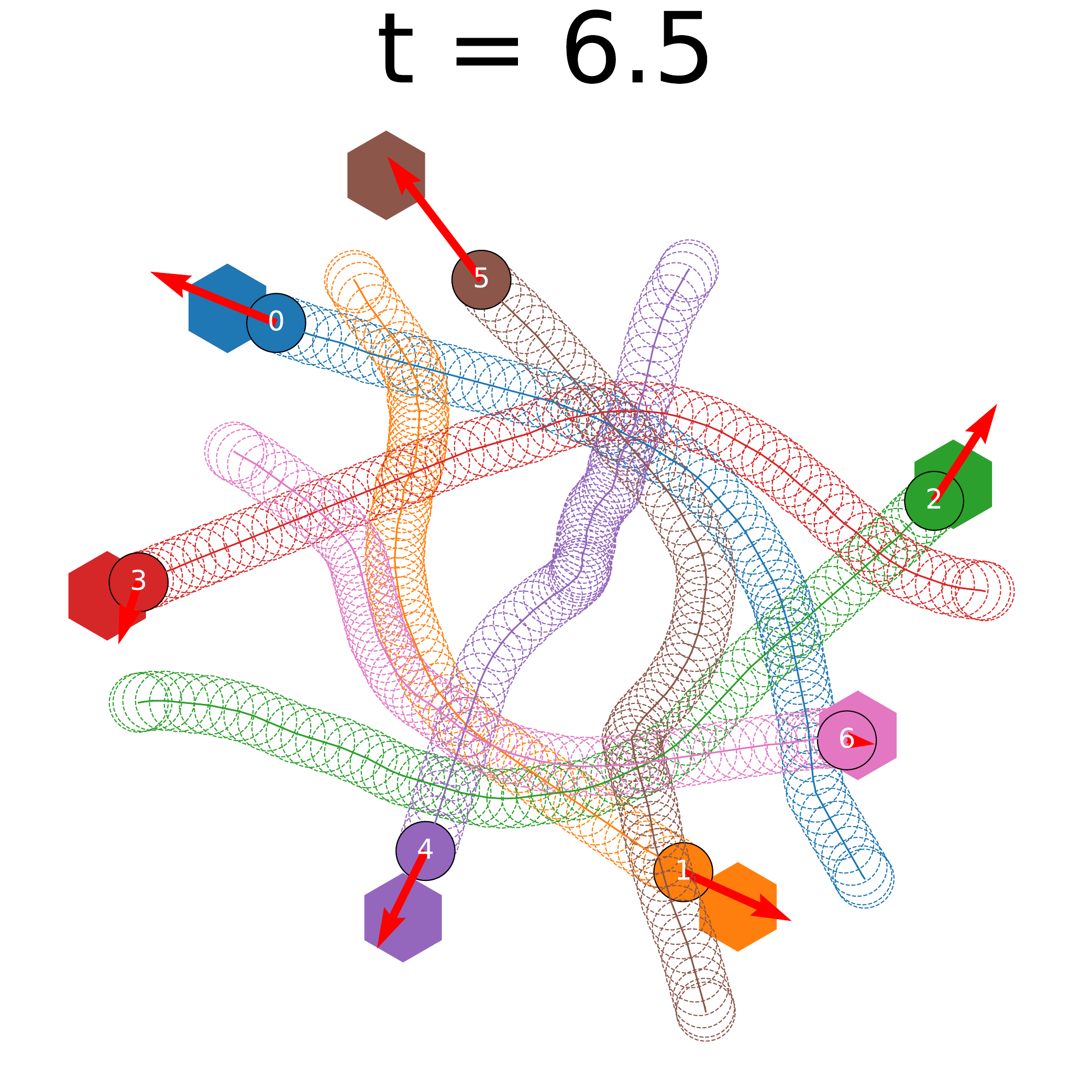}%
    \hfill
    \includegraphics[width=0.15\textwidth]{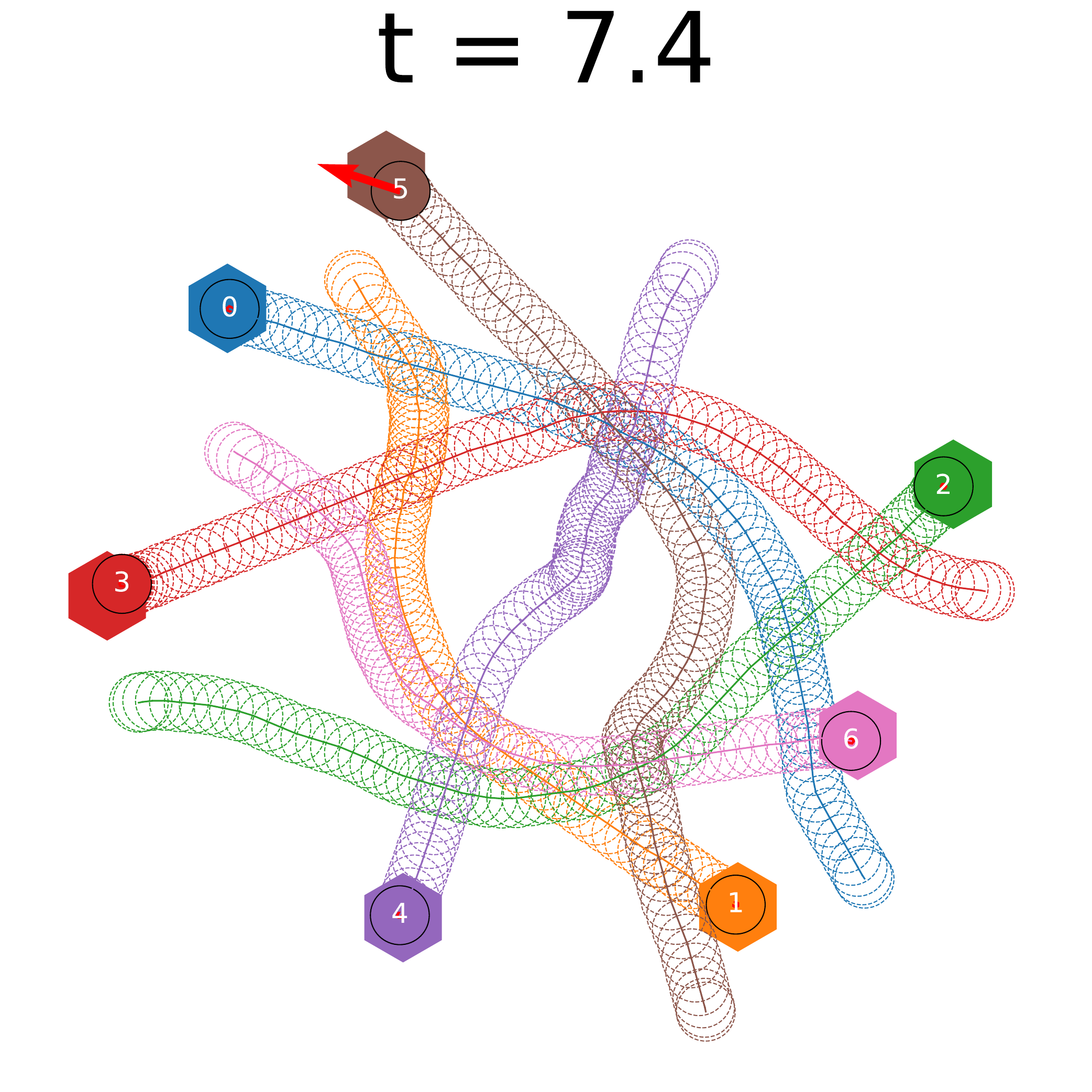}%
    \label{fig:traj_compare:wnum}
  }
  \caption{
    % Comparison of the temporal evolution of trajectories generated by T-MPC~\cite{Mav23} and WNumMPC (ours) in the same Crossing scenario with $N=7$ agents. 
    % (a) T-MPC repeatedly switches between passing decisions, so that all agents except the red and green ones exhibit oscillatory behavior and remain around the same locations at $t=1.5$, $2.5$, and $3.5$, which delays the overall resolution. 
    % (b) WNumMPC breaks the symmetry through the learned signed target winding numbers and weights and enables continuous navigation without unnecessary stopping for all agents except the purple one, while the purple agent temporarily halts because it becomes surrounded by neighboring agents, which constitutes an appropriate yielding behavior.
    Temporal evolution of trajectories in the same Crossing scenario with $7$ agents. 
    (a) T-MPC~\cite{Mav23} shows oscillatory switching and prolonged stagnation except for the red and green agents (e.g., $t=1.5, 2.5, 3.5$). 
    (b) WNumMPC (ours) breaks symmetry using signed target winding numbers and weights, enabling continuous navigation except when the purple agent temporarily gives way to surrounding agents.
  }
  \label{fig:traj_compare}
\end{figure*}

\subsection{Real-World Results}
\cref{success_rate,ave_etg} (right) show the real-world results and, for reference, the corresponding differential-drive simulation.
%WNumMPC achieved the highest success rate on both Random and Crossing. On Crossing, it also attained lower extra time than T-MPC, which had the second-highest success rate.
Compared to the holonomic simulation results, the performance gaps between methods became smaller. Therefore, in addition to reporting the empirical metrics, we examined whether the observed differences are statistically significant by applying hypothesis tests. Specifically, we conducted McNemar's test for success rates and the Wilcoxon signed-rank test for extra time to goals\footnote{For each method pair, we compared the results of instances where both succeeded.}, both implemented in SciPy~\cite{virtanen2020scipy}.

% On Crossing instances, WNumMPC achieved the significantly higher success rate compared to other methods.
% We conducted McNemar's test using SciPy~\cite{virtanen2020scipy} and verified that WNumMPC has higher success rates than Vanilla MPC with $p=5\times 10^{-7}$ and than T-MPC with $p=0.02$.
On Crossing instances, WNumMPC achieved a significantly higher success rate compared to other methods. 
We verified that WNumMPC achieved higher success rates in real-world experiments than Vanilla MPC with $p=5\times 10^{-7}$ and than T-MPC with $p=0.02$. 
%While vanilla MPC had a smaller average extra time, it is the average only for successful instances. When averaging only the successful instances for both Vanilla MPC and the proposed method, the proposed method yielded a slightly better value (Vanilla MPC: $4.309$\,s, WNumMPC: $4.301$\,s). The same holds true when making a similar comparison with T-MPC (T-MPC: $4.59$\,s, WNumMPC: $4.46$\,s).
The same trend is observed in the differential-drive simulation, where WNumMPC also demonstrated a higher success rate in simulation than Vanilla MPC with $p=0.005$ and than T-MPC with $p=0.01$. These results support the advantage of our approach in symmetric scenarios where agents must break the symmetry of how to pass each other.
% Our method demonstrated the significantly higher success rate also in simulation than Vanilla MPC with $p=0.005$ and than T-MPC with $p=0.01$
% Compared to the results in simulation, both the existing and proposed methods showed a tendency for performance degradation in real-world experiments in these instances and it is observed that the performance gap between simulation and real-world was higher in Vanilla MPC than in T-MPC and in WNumMPC. Indeed, while the success rates of vanilla MPC and of T-MPC are almost equal, vanilla MPC has significantly ($p=0.004$) low success rate than T-MPC in real world. 
% Considering winding numbers appears to enhance the robustness.
%On Random, while WNumMPC’s success rate was high, its extra time was slightly worse. Because a similar tendency was observed in simulation, it is likely due to kinematic differences. For differential-drive robots, safety-driven evasive actions may cause larger time penalties than for holonomic agents. 
On Random instances, WNumMPC also attained the highest success rate, while we did not observe statistically significant differences in success rates between methods.
Extra time to goal showed no significant differences between methods on both Crossing and Random instances in real-world experiments.
This is likely because, in holonomic scenarios, even when deadlocks occurred, they often resolved gradually over time, whereas in non-holonomic scenarios, such situations resulted in timeouts or collisions and were not factored into the average calculation.

% More importantly, although WNumMPC relies on a learned planner and is therefore expected to be more susceptible to the simulation-to-real gap than other baselines, we found that explicitly leveraging winding numbers mitigates this effect. In particular, WNumMPC exhibited the smallest simulation-to-real degradation in success rate across both scenarios, suggesting that winding-number-aware policy improves robustness when transferring policies from simulation to real robots.
%Moreover, although WNumMPC relies on a learned planner and is therefore expected to be more susceptible to the simulation-to-real gap than other baselines, we found that explicitly leveraging winding numbers mitigates this effect.
%ompared with Vanilla MPC (Crossing: $21\%$, Random: $6\%$), both WNumMPC (Crossing: $8\%$, Random: $1\%$) and T-MPC (Crossing: $8.5\%$, Random: $4\%$) exhibited substantially smaller simulation-to-real degradations in success rate.
Moreover, comparing the success rate differences between simulation and real-world scenarios, we found that both WNumMPC (Crossing: $8\%$, Random: $1\%$) and T-MPC (Crossing: $8.5\%$, Random: $4\%$) exhibited substantially smaller degradations than Vanilla MPC (Crossing: $21\%$, Random: $6\%$).
This fact suggests that exploiting winding-number-aware strategies improves robustness when transferring policies from simulation to real robots.
It is noteworthy that WNumMPC achieves both the advantages of learning-based approaches and robustness.

% In summary, we observed that WNumMPC is effective also in non-holonomic scenarios, and that incorporating winding numbers enhances robustness against the simulation-to-real gap.
%% ページ数に余裕がなさそうなので余分な文章は極力省く
% These results demonstrate that the learned topological cooperative strategy is effective in the real world,
% particularly in symmetric situations prone to deadlocks.
% Notably, the ability to effectively avoid the deadlocks that were the primary cause of performance degradation 
% for existing methods in the Crossing scenario suggests the superiority of our cooperative strategy learning approach
% in real-world navigation tasks requiring complex interactions and symmetry breaking. 
% On the other hand, it also became apparent that the learning-based Planner is susceptible to the Sim-to-Real gap.
% Bridging this gap to further enhance the method's reliability will be an important task for future research.

\subsection{Qualitative Evaluation in Simulation Experiments}
In addition to the quantitative comparisons above, we visualized and qualitatively analyzed the results of our simulation experiments.

\subsubsection{Comparison of Trajectories} \cref{fig:trajectories,fig:traj_compare} provide a qualitative evaluation by visualizing representative trajectories and their time evolution in the same Crossing instance.
\cref{fig:trajectories} visualizes trajectories for the same Crossing instance. 
In ORCA, agents stopped after yielding to each other. 
In CADRL, agents failed to avoid collisions after gathering in the center. 
Vanilla MPC successfully avoided collisions, but many agents temporarily stopped. 
T-MPC avoided collisions, but exhibited temporary stops, albeit less pronounced than Vanilla MPC, and some agents took long detours. 
In contrast, WNumMPC successfully avoided collisions efficiently.

\cref{fig:traj_compare} illustrates the temporal evolution of trajectories for T-MPC and WNumMPC in a Crossing scenario with $N=7$ agents.
In T-MPC (\cref{fig:traj_compare:tmpc}), due to the strong symmetry of the scenario, agents struggle to resolve conflicts.
Specifically, at $t=1.5, 2.5,$ and $3.5$, all agents except the Red and Green ones fail to decide on a passing direction, exhibiting oscillatory behavior and remaining in the same location.
In contrast, WNumMPC (\cref{fig:traj_compare:wnum}) effectively breaks symmetry using the learned topological strategy.
All agents, except the Purple one, navigate continuously without unnecessary stops.
The Purple agent halts temporarily, but this is due to being surrounded by other agents, which suggests a reasonable yielding behavior to avoid collisions.

% Based on the above results, the effectiveness of the proposed method and its superiority over existing methods are demonstrated.
% These results suggest that while existing methods utilize either learning to handle complex interactions or optimization to achieve efficient navigation and safety, the proposed method employs both approaches. This enables it to effectively address both safety and efficiency.

\subsubsection{Analysis of Learned Topological Strategies}
\begin{figure*}[t]
  \centering
  \subfloat[
    The transition of the weight~$\boldsymbol{\alpha}^0_k$ output by the policy.
  ]{
    \includegraphics[width=0.19\textwidth]{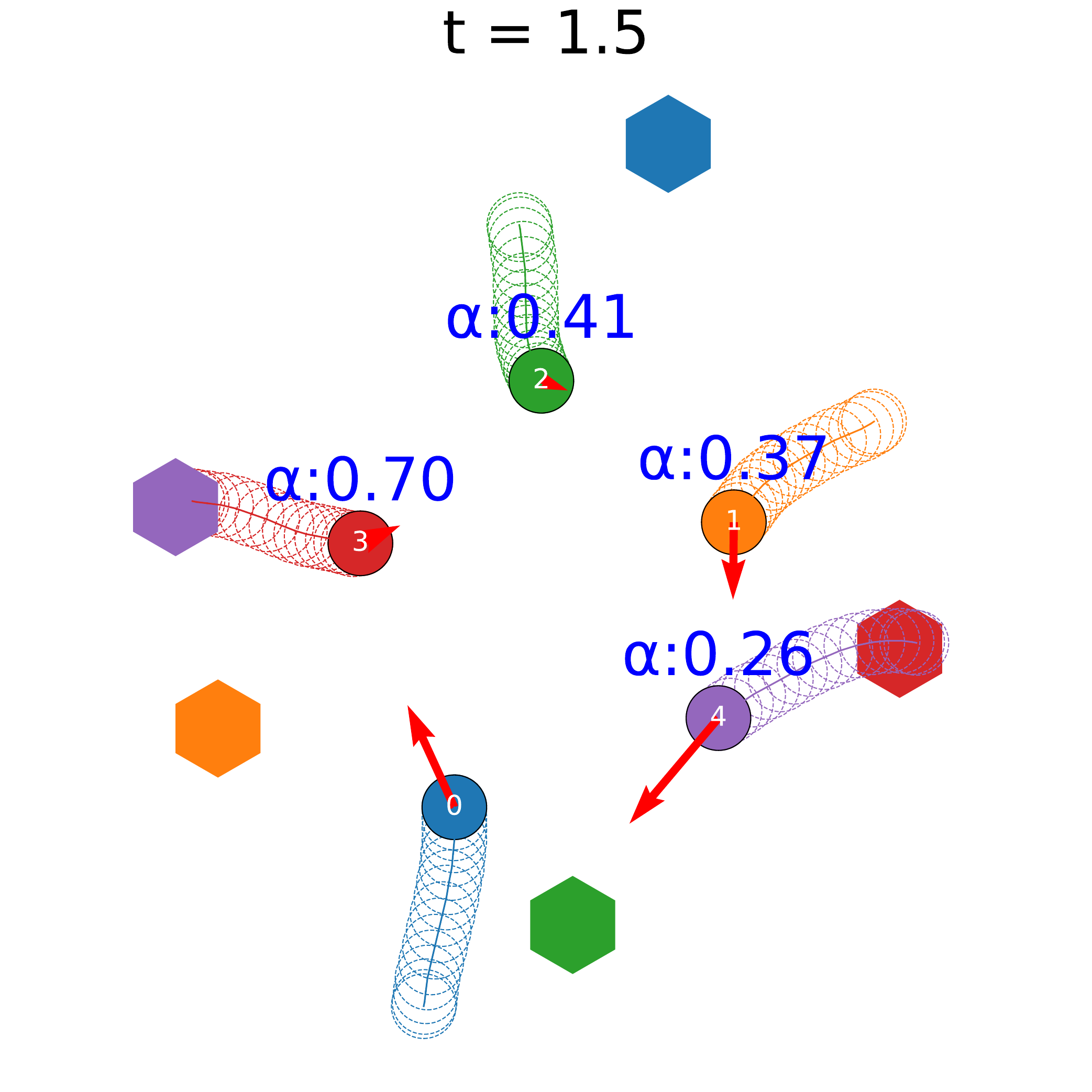}%
    \qquad \quad
    % \includegraphics[width=0.2\textwidth]{fig_iros/weight_plot/wnum_mpc/h4_opp_ep5/step25.png}%
    % \hfill
    % \includegraphics[width=0.25\textwidth]{fig_iros/weight_plot/wnum_mpc/h4_opp_ep5/step35.png}%
    \includegraphics[width=0.19\textwidth]{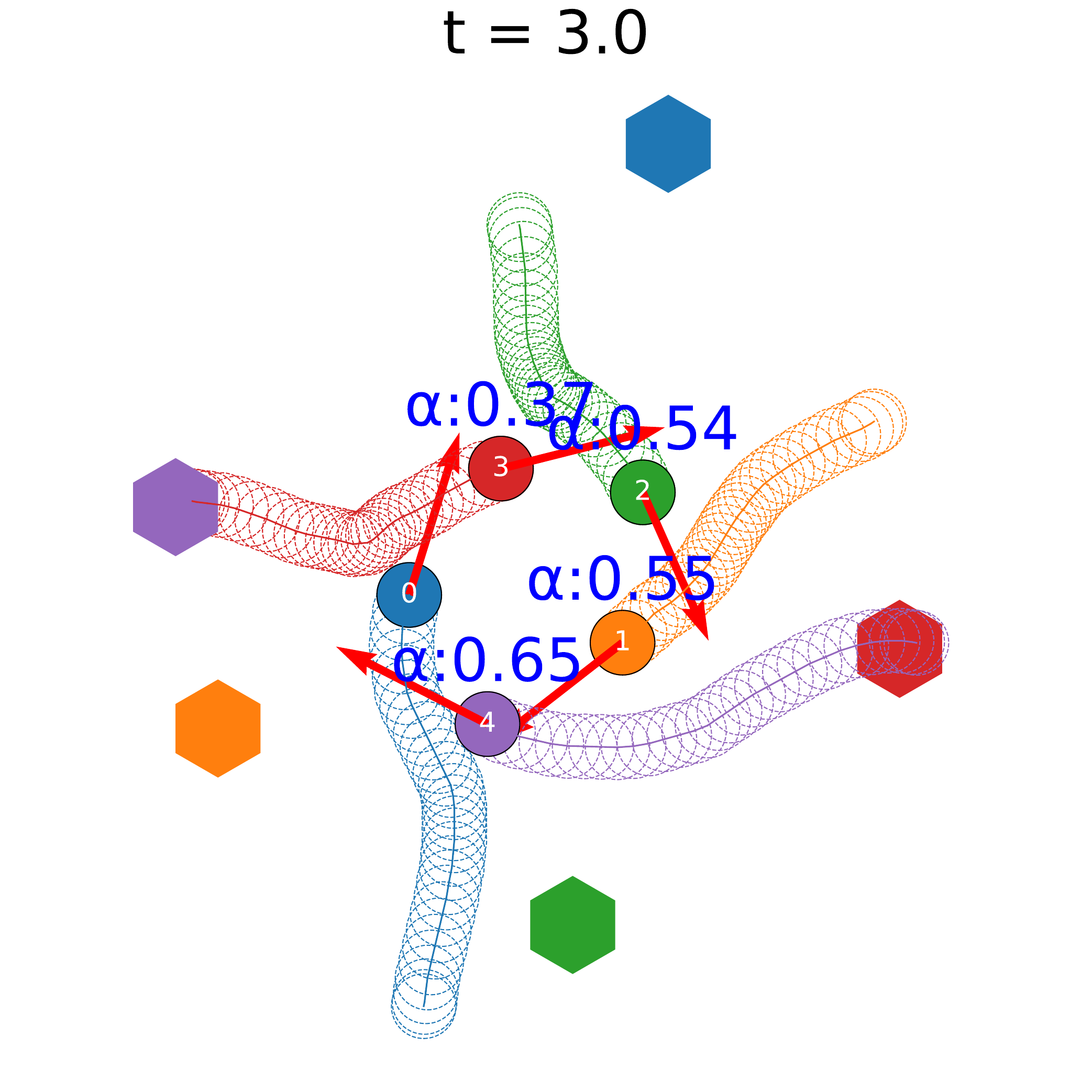}%
    \label{fig:weights:alpha}
  }
  \qquad \quad
  \subfloat[
    The transition of the winding number~$\bw^0_k$ output by the policy.
  ]{
    \includegraphics[width=0.19\textwidth]{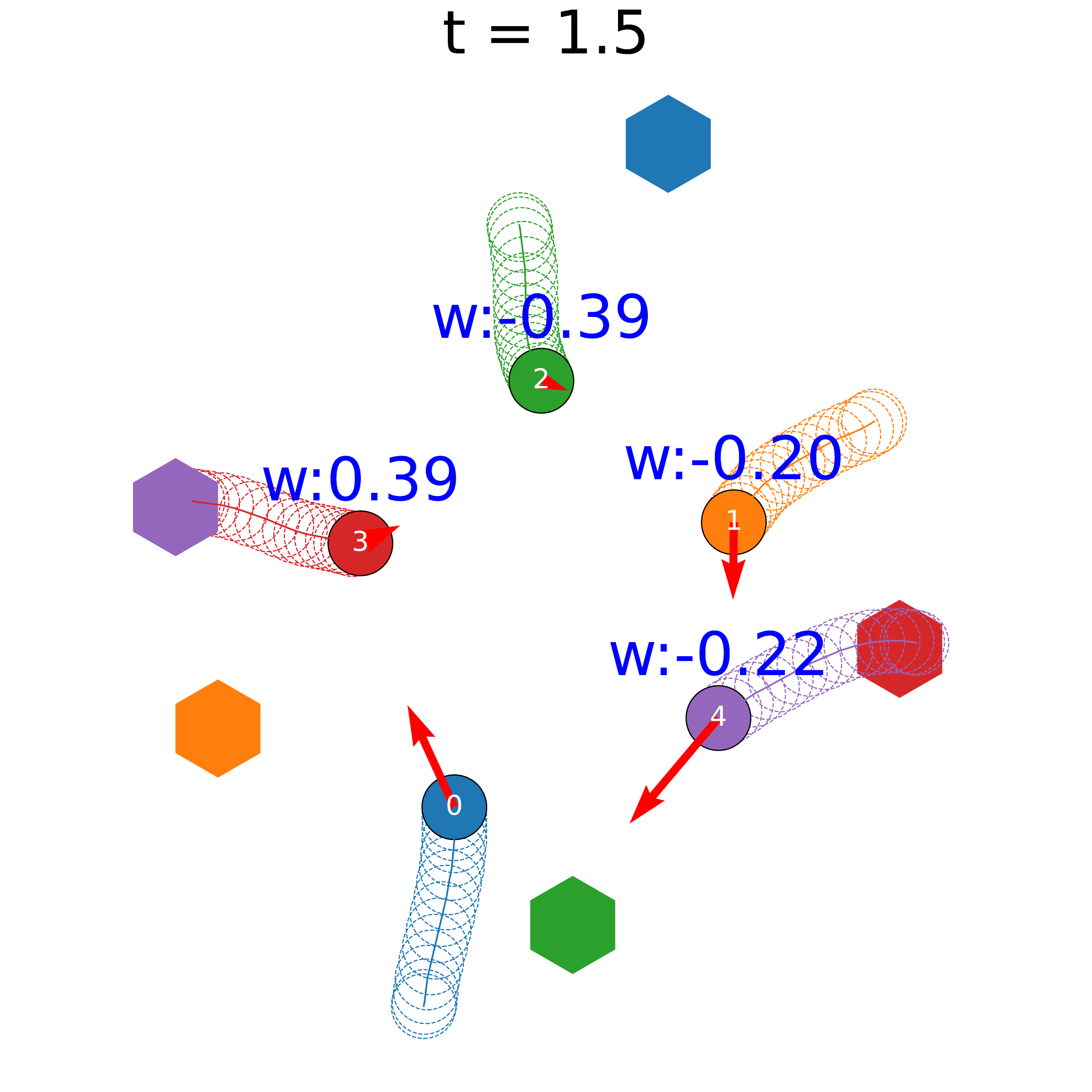}%
    \qquad \quad
    \includegraphics[width=0.19\textwidth]{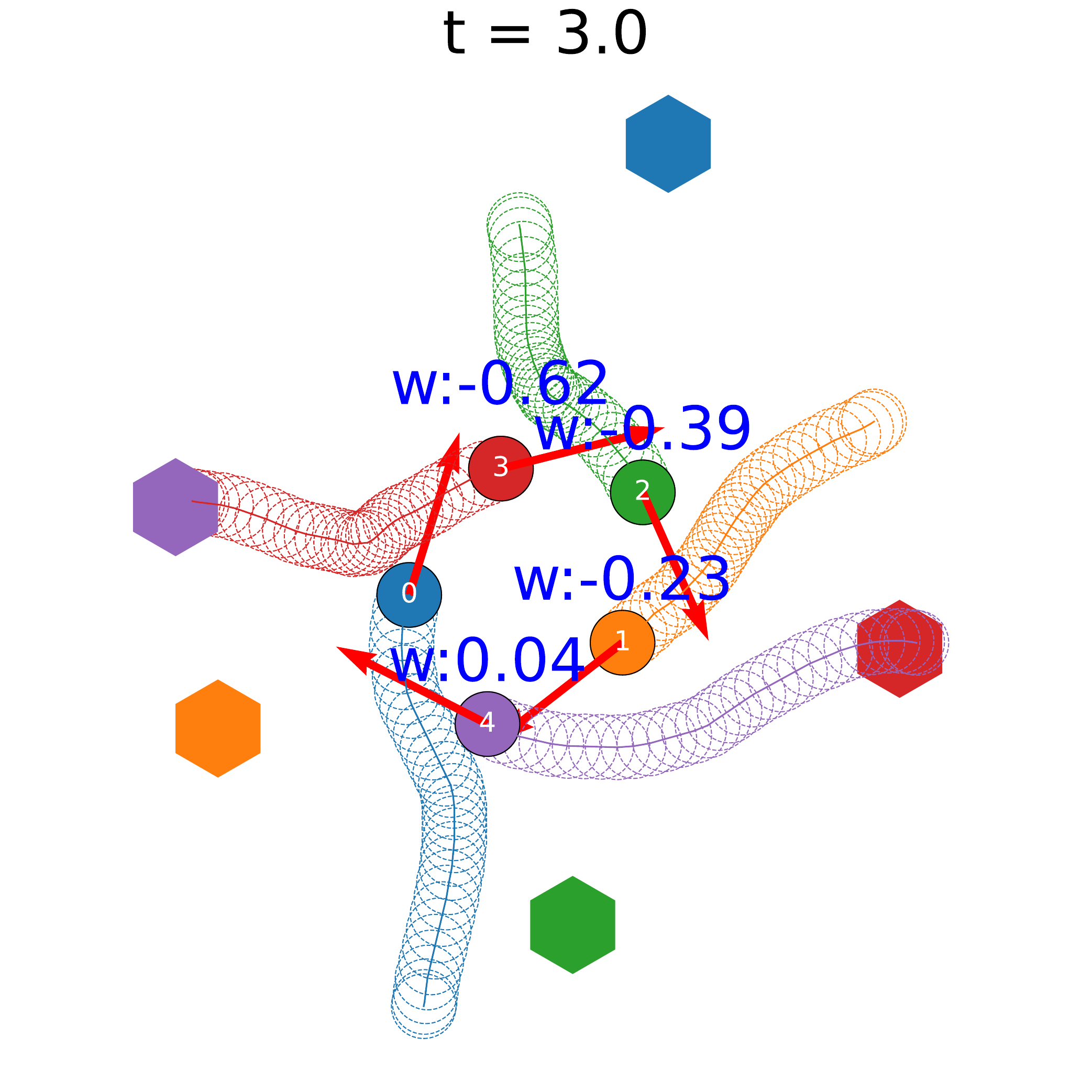}%
    \label{fig:weights:w}
  }
  \caption{
    % Example of the Planner outputs of agent~$0$ in WNumMPC during a Crossing scenario with $N=5$ agents.
    % (a) Early in the episode ($t=1.5$) the Planner assigns a large weight to the red agent, while at $t=3.0$ it spreads weights across the remaining agents as the passing side for the red agent becomes effectively determined, reducing the need for an additional topological constraint. 
    % (b) Early in the episode ($t=1.5$) the Planner plans left-avoidance for all agents except the red one (right-avoidance), but as the actual crossing time approaches ($t=3.0$) it shifts toward a policy consistent with left-avoidance. And, in the $t=3.0$, the purple agent is assigned $w \approx 0$, indicating a strategy that avoids inducing relative rotation with the purple agent.
    Example Planner outputs of agent~$0$ (blue) in WNumMPC for a Crossing scenario with $5$ agents.
    (a) Early in the episode ($t=1.5$) the Planner assigns a large weight to the red agent, while it spreads weights across the remaining agents as the passing side for the red agent becomes effectively determined ($t=3.0$).
    (b) Early in the episode ($t=1.5$) the Planner plans left-avoidance for all agents except the red one (right-avoidance), but as the actual crossing time approaches ($t=3.0$) it shifts toward a policy consistent with left-avoidance. 
  }
  \label{fig:weights}
\end{figure*}

To validate the internal decision-making process of the proposed method, we analyze the temporal evolution of the Planner's outputs ($w_{k}^{i}$ and $\alpha_{w,k}^{i}$) in a Crossing scenario with $N=5$ agents, as visualized in \cref{fig:weights}.

\cref{fig:weights:alpha} illustrates the transition of the dynamic weights $\boldsymbol{\alpha}_{w,k}^i$. In the early phase of the episode ($t=1.5$), the Planner assigns a large weight to the Red agent ($\alpha=0.70$), effectively identifying it as the most critical peer for immediate collision avoidance. At $t=3.0$, the weights are distributed more evenly among the agents currently crossing. The weight for the Red agent decreases ($\alpha=0.37$) as the interaction is nearly resolved. Notably, the Purple agent retains a high weight ($\alpha=0.65$); this can be interpreted as the Planner prioritizing safety against the Purple agent to prevent a potential rear-end collision from behind.

\cref{fig:weights:w} shows the behavior of the target winding numbers $\mathbf{w}_{k}^i$. At $t=1.5$, the Planner explicitly differentiates strategies, planning a right-avoidance for the red agent while suggesting left-avoidance for the others. As time progresses to the actual crossing moment ($t=3.0$), the policy converges to a consistent left-avoidance plan for the conflicting agents. Furthermore, the target winding number for the Purple agent at $t=3.0$ is set close to zero ($w \approx 0.04$). 
This indicates that the Planner intends to maintain the current relative angle without inducing unnecessarily large rotation, thereby achieving a stable avoidance.

%%%%%%%%%%%%%%%%%%%%%%%%%%%%%%%%%%%%%%%%%%%%%%%%%%%%%%%%%%%%%%%%%%%%%%%%%%%%%%%%
%%%%%%%%%%%%%%%%%%%%%%%%%%%%%%%%%%%%%%%%%%%%%%%%%%%%%%%%%%%%%%%%%%%%%%%%%%%%%%%%

\section{Conclusion}
\label{sec:conclusion}
In this work, we proposed a novel hierarchical control framework that learns topological cooperative strategies 
to address deadlocks arising from symmetries between agents in decentralized multi-agent navigation. 
% This method consists of a learning-based planner that devises topological cooperative strategies 
% and prioritizes coordination among multiple agents,
% and a model-based controller that generates safe and efficient trajectories based on the plan.
% Through experiments in both simulation and on real hardware,
% we demonstrated that the proposed method outperformed existing approaches in terms of both success rate and efficiency,
% particularly in high-density scenarios. These results suggest that our approach of learning cooperative strategies themselves,
% rather than relying on fixed rules, is effective for efficiently breaking symmetries between agents and achieving efficient navigation.
Through experiments in both simulation and real-world settings, we demonstrated that the proposed method achieves consistently strong performance even in dense, symmetry-prone scenarios. 
These results imply that our approach of learning cooperative strategies themselves is effective for breaking symmetries between agents, achieving efficient navigation and minimizing simulation-to-real performance degradation.
%Moreover, our method exhibited the smallest simulation-to-real degradation in success rate. This result suggests that explicitly leveraging winding numbers improves robustness when transferring policies from simulation to real robots.

% As for future work, we envision three directions based on the results of this study.
% The first is to improve the robustness and generalization of the planner.
% By incorporating sim-to-real techniques such as Domain Randomization into the planner's training process,
% we aim to obtain a highly reliable policy that can bridge the sim-to-real gap.
% The second direction is to expand the scope of application by advancing the controller.
% Introducing a state-of-the-art nonlinear model predictive control, such as Model Predictive Path Integral (MPPI), 
% into the controller will enable the application of our framework to agents with more complex dynamics and enhance its control performance. 
As for future work, we envision three directions. First, by predicting current states of other agents from observations rather than being given them directly, we can extend our approach to more realistic situations. Second, introducing state-of-the-art nonlinear model predictive control into the Controller will enable the application of our framework to agents with more complex dynamics and enhance control performance. Third, incorporating Graph Neural Networks into the Planner is expected to improve scalability to large-scale agent groups and enhance the generality of the proposed method, such as adaptability to scenarios with agent counts different from those in training.

\bibliographystyle{IEEEtran}
\bibliography{IEEEabrv,refs}

\end{document}